\documentclass{article}
\usepackage[preprint]{neurips_2024}
\usepackage[utf8]{inputenc} 
\usepackage[T1]{fontenc}    
\usepackage{hyperref}       
\usepackage{url}            
\usepackage{booktabs}       
\usepackage{amsfonts}       
\usepackage{nicefrac}       
\usepackage{microtype}      
\usepackage{xcolor}         
\usepackage{amsmath,amssymb,amsfonts}
\usepackage{graphicx}
\usepackage{algorithmic}
\usepackage[linesnumbered,ruled,vlined, ruled,vlined]{algorithm2e}
\usepackage{amsmath}
\usepackage{tikz}
\usepackage{lineno}
\usepackage{pgfplots}
\pgfplotsset{compat=1.18}
\usepackage{subcaption}
\usepgfplotslibrary{fillbetween}
\usepackage{float}
\usepackage{multirow}
\usetikzlibrary{shapes.geometric, arrows}
\usepackage{circuitikz}
\usepackage{booktabs}
\usepackage[inkscapelatex=false]{svg}
\title{Conformalized Prediction of Post-Fault Voltage Trajectories Using Pre-trained and Finetuned Attention-Driven Neural Operators}
\author{%
  Amirhossein~Mollaali\\
  School of Mechanical Engineering\\
  Purdue University\\
  West Lafayette, IN, USA, 47906\\
  \texttt{amollaal@purdue.edu} \\
  \And
  Gabriel~Zufferey\\
  Facultad de Ingeniería Eléctrica y Electrónica\\
  Escuela Politécnica Nacional\\
  Quito, Ecuador, 170525\\
  \texttt{gabriel.zufferey@epn.edu.ec} \\
  \AND
  Gonzalo~Constante-Flores\\
  Davidson School of Chemical Engineering\\
  Purdue University\\
  West Lafayette, IN, USA, 47906\\
  \texttt{geconsta@purdue.edu}\\
  \And
  Christian~Moya\\
  Department of Mathematics\\
  Purdue University\\
  West Lafayette, IN, USA, 47906\\
  \texttt{cmoyacal@purdue.edu} \\
  \AND
  Can~Li\\
  Davidson School of Chemical Engineering\\
  Purdue University\\
  West Lafayette, IN, USA, 47906\\
  \texttt{canli@purdue.edu}\\
  \And
  Guang Lin\\
  Department of Mathematics and School of Mechanical Engineering\\
  Purdue University\\
  West Lafayette, IN, USA, 47906\\
  \texttt{guanglin@purdue.edu} \\
  \And 
  Meng Yue\\
  Interdisciplinary Science Department\\
  Brookhaven National Laboratory\\
  Upton, NY, USA, 11973\\
  \texttt{yuemeng@bnl.gov}
}
\begin{document}
\maketitle
\begin{abstract}
  This paper proposes a new data-driven methodology for predicting intervals of post-fault voltage trajectories in power systems. We begin by introducing the Quantile Attention-Fourier Deep Operator Network (QAF-DeepONet), designed to capture the complex dynamics of voltage trajectories and reliably estimate quantiles of the target trajectory without any distributional assumptions. The proposed operator regression model maps the observed portion of the voltage trajectory to its unobserved post-fault trajectory. Our methodology employs a pre-training and fine-tuning process to address the challenge of limited data availability. To ensure data privacy in learning the pre-trained model, we use merging via federated learning with data from neighboring buses, enabling the model to learn the underlying voltage dynamics from such buses without directly sharing their data. After pre-training, we fine-tune the model with data from the target bus, allowing it to adapt to unique dynamics and operating conditions. Finally, we integrate conformal prediction into the fine-tuned model to ensure coverage guarantees for the predicted intervals. We evaluated the performance of the proposed methodology using the New England 39-bus test system considering detailed models of voltage and frequency controllers. Two metrics, Prediction Interval Coverage Probability (PICP) and Prediction Interval Normalized Average Width (PINAW), are used to numerically assess the model's performance in predicting intervals. The results show that the proposed approach offers practical and reliable uncertainty quantification in predicting the interval of post-fault voltage trajectories.
\end{abstract}

\section{Introduction} \label{sec:introduction}
Stability in power systems refers to the system's inherent ability to return to an operational equilibrium following a physical disturbance, such as a contingency \citep{Davidhill2021, Shair2021}. Assessing this ability is essential for ensuring the continuous and reliable operation of power systems. To maintain stability after potential contingencies like faults or equipment failures, system operators conduct offline dynamic studies in a preventive manner \citep{VanCutsem2000}. These studies evaluate the system's dynamic response to faults and disconnections of components by using model-based simulators, which numerically solve a set of nonlinear differential-algebraic equations (DAEs) to track system trajectories after each contingency \citep{Andersonbook, Taha2021, Taha2022}. Although model-based approaches provide highly accurate evaluations, their significant computational costs make them impractical for real-time applications, limiting their usefulness in situations that require rapid decision-making \citep{Morison2004, Zhao2013, Yousefian2017}.

To overcome these challenges, modern measurement systems—such as phasor measurement units (PMUs) and advanced communication technologies—now provide operators with real-time data that can enhance stability analysis and help secure the electrical energy supply \citep{Vladimir2011, Paul2018, Zhao2019}. In this context, data-driven methodologies have emerged as a promising alternative to traditional physics-based methodologies. These approaches use historical or synthetic data to predict power system stability with significantly faster computation times. Our work advances this class of data-driven methods to improve the effectiveness and efficiency of stability analysis in power systems.

In recent years, machine learning (ML) techniques have been employed to assess various metrics of power system stability by leveraging real-time data from PMUs, enabling the development of real-time stability assessment tools. Some of the most prominent ML models include artificial neural networks \citep{Nima2007}, convolutional neural networks \citep{Zhongtuo2020}, ensemble learning \citep{liu_data-driven_2020}, stacked autoencoders \citep{yin_deep_2018}, and long short-term memory networks \citep{Zonghe2024}. However, traditional deep learning methods operate in finite-dimensional spaces, limiting their ability to handle the complexity of power system dynamics. Voltage trajectories are inherently infinite-dimensional as continuous functions of time and other parameters. By accounting for these infinite dimensions, function spaces offer a more accurate representation of these trajectories compared to finite-dimensional vector spaces.

Operator learning has emerged as a powerful framework for mapping function spaces \citep{lu2021learning, TRIPURA2023115783, li2020fourier, LU2022114778, zhu2023fourier}. A prominent method of operator learning is the Deep Operator Network (DeepONet) \citep{lu2021learning, jin2022mionet, HOWARD2023112462, zhang2022belnet, lin2023learning, moya2023approximating}. DeepONet architecture efficiently addresses the complexities of function space mappings, substantially impacting scientific machine learning by providing robust tools for solving high-dimensional nonlinear problems. DeepONet has shown notable success in advanced mathematical applications, such as approximating the solution operator of parametric partial differential equations (PDEs) \citep{jiao2024solving, zhang2024bayesian, wang2021learning, LIN2023111713, hao2023pinnacle} and multi-scale problems \citep{zhang2024bayesian, lin2021operator}. Beyond theoretical achievements, the practical utility of DeepONet is demonstrated in real-world applications, such as optimizing the rib profile \citep{SAHIN2024124813}, predicting the solution of disk–planet interactions in protoplanetary disks \citep{Mao_2023}, forecasting stresses in elastoplastic structures \citep{HE2023116277}, and predicting the crack path \citep{GOSWAMI2022114587}.

DeepONet has also shown significant potential in power systems, particularly for predicting post-fault trajectories. In this context, in \citep{MOYA2023166} authors proposed a method utilizing DeepONet-based models to estimate predictive intervals for these trajectories. They first employed B-DeepONet, which uses stochastic gradient Hamiltonian Monte Carlo for uncertainty quantification but requires numerous hyperparameters, making the training process expensive. To address this, the authors introduced Prob-DeepONet, which offers automated uncertainty quantification with minimal computational overhead through a probabilistic training strategy. However, this approach assumes a normal distribution for the target solution, which is not necessarily accurate and can negatively affect the prediction performance. Additionally, their methodology is unable to guarantee the coverage of the predicted intervals. On the other hand, their method only uses data from the target bus for training the model and cannot utilize data from neighboring buses, making it prone to challenges of low-data regimes in power systems. To overcome the limitations of low-data regimes,  the authors of \citep{10210333} proposed DeepGraphNet to predict post-fault trajectories. Their architecture incorporates graph neural networks, enabling the framework to learn from data across neighbor buses in the power system, thus mitigating low-data regime issues. However, they used a centralized learning approach to train the model, which violates the privacy of the buses, making its practical implementation challenging. Additionally, DeepGraphNet is deterministic, lacking the capability to provide uncertainty quantification for its predictions. Moreover, the models proposed in these studies lack the ability to handle heterogeneous input data across different time domains, which reduces their practicality for power systems, considering the randomness of fault occurrences in these systems.

Existing approaches do not offer a comprehensive solution for predicting post-fault trajectory intervals without relying on distributional assumptions. Additionally, these models are unable to guarantee the coverage probability of the predicted intervals. Furthermore, they fail to address the challenges of low-data regimes while maintaining the privacy of individual buses.

To bridge this gap, we propose a comprehensive approach for predicting the intervals of post-fault voltage trajectories in power systems. Our methodology focuses on the development of a quantile-based operator learning model that estimates predictive intervals without relying on distributional assumptions, making it more flexible and robust across different scenarios. To overcome issues related to limited data, we implement a pre-training and fine-tuning strategy. In the pre-training phase, we use data from neighboring buses to capture the underlying dynamics of voltage trajectories. This phase incorporates federated learning to maintain data privacy while enhancing the model's robustness. After pre-training, the model is fine-tuned with data from the target bus, which further improves the accuracy and reliability of predictions for that specific bus. Finally, to ensure that the predicted intervals have the desired coverage guarantees, we integrate conformal prediction methods into the fine-tuning process.

The main contributions of this work are as follows:

\begin{itemize}

    \item \textit{QAF-DeepONet:} We introduce QAF-DeepONet, a novel data-driven framework for predicting post-fault voltage trajectory intervals. It employs an attention mechanism and a Fourier feature network to capture dynamic behaviors and fluctuations in voltage trajectories while handling heterogeneous input data. QAF-DeepONet estimates lower and upper quantiles of the target solution to capture the range of possible outcomes.
    
    \item \textit{Federated Learning Pre-Training:} We employ federated learning to pre-train QAF-DeepONet using distributed data from neighboring buses. This approach leverages the available data to address the issue of low-data regime and to enhance the model's generalizability and robustness via merging while preserving data privacy.
    
    \item \textit{Fine-Tuning for Targeted Bus:} We fine-tune the pre-trained model with data from the target bus, improving the reliability and precision of the estimated predictive intervals.
    
    \item \textit{Conformal Prediction Integration:} We integrate conformal prediction techniques into the fine-tuned QAF-DeepONet predictions, ensuring the predicted intervals meet the desired coverage guarantees. This adjustment provides statistically reliable intervals, supporting effective decision-making in power systems.
\end{itemize}

The remainder of the paper is organized as follows: Section~\ref{sec:Problem_Definition} introduces the problem statement and the main challenges of predicting post-fault trajectories. Section~\ref{sec:methodology} presents the proposed methodology and the training strategy. Section~\ref{sec:Numerical_Experiments} shows the numerical experiments to evaluate the performance of the proposed methodology. Finally, Sections\ref{sec:future-work} and~\ref{sec:conclusion} describe out future work and conclude the paper.
\section{Preliminaries} \label{sec:Problem_Definition}
In this section, we outline the problem statement and the primary challenges associated with predicting post-fault trajectories. 

\subsection{Problem Statement}
Assessing the short-term stability of a power system is imperative to ensuring the continuous and reliable operation of power systems. This type of stability refers to the power system's ability to find a stable equilibrium point after being subject to disturbances, such as faults or equipment failures. The dynamic behavior of the system's trajectories is governed by: (i) the dynamics of mechanical (e.g., turbines), electrical (e.g., generators, lines, transformers, loads), and control systems (e.g., governor, and automatic voltage regulators) and (ii) the algebraic physical laws determining the power flows throughout the power grid. Such behavior can be modeled as a system of DAEs across three different stages: pre-fault, fault-on, and post-fault. Each stage has its own unique set of dynamics that determine the stability and transient behavior of the system's trajectories.

The behavior of the system's trajectories, illustrated in Figure~\ref{Picture1}, during these stages, can be analyzed by examining them as follows:

\begin{enumerate}
    \item \textit{Pre-Fault Dynamics}: 
    Prior to any disturbance, the power grid is assumed to be in an equilibrium state at time \( t_0 \). This equilibrium persists as long as the system remains undisturbed, up until a fault occurs at time \( t_f \). During this pre-fault stage, the dynamics of the system are governed by the following set of DAEs:
    \begin{subequations}
    \begin{align}
        \dot{x}(t) &= f_{P}(x(t), z(t)), \quad \text{for } t \in [t_0, t_f], \\
        0 &= g_{P}(x(t), z(t)),
    \end{align}
    \end{subequations}
    where \( x(t) \) represents the state vector of the system, and \( z(t) \) is the set of algebraic variables. These equations describe the stable operation of the system prior to any fault, assuming the system is initially in a steady state \citep{Sauerbook}.
    
    \item \textit{Fault-on Dynamics}: 
    When a fault occurs at time \( t_f \), the system transitions into a faulted condition, which leads to a temporary topological change in the system. This faulted state lasts until the fault is cleared at time \( t_{cl} \) by a protective relay, which isolates the faulted component. The time interval from the occurrence of the fault to its isolation is referred to as clearing time. During this period, the dynamics of the system are altered and are described by the following set of DAEs:
    \begin{subequations}
    \begin{align}
        \dot{x}(t) &= f_{F}(x(t), z(t)), \quad \text{for } t \in [t_f, t_{cl}], \\
        0 &= g_{F}(x(t), z(t)).
    \end{align}
    \end{subequations}
    The fault-on dynamics capture the system's immediate response to the disturbance, reflecting the changes in system behavior caused by the fault.

    \item \textit{Post-Fault Dynamics}: 
    After the fault is cleared at time \( t_{cl} \), the system enters a transient response as it attempts to return to a stable equilibrium. This phase continues until a specified time horizon, \( T \), and the dynamics during this stage are governed by:
    \begin{subequations}
    \begin{align}
        \dot{x}(t) &= f_{PF}(x(t), z(t)), \quad \text{for } t \in [t_{cl}, T], \\
        0 &= g_{PF}(x(t), z(t)).
    \end{align}
    \end{subequations}
    The post-fault dynamics are key for understanding the system's ability to regain stability after a disturbance and to identify potential instabilities that may arise after the fault occurs.
\end{enumerate}

\subsection{Problem Setup}
This paper focuses on the real-time prediction of post-fault voltage trajectory intervals using local measurements. The approach is structured into two parts: offline training and online inference. In the offline training stage, the model is trained using historical data from the target bus and its neighboring buses. During online inference, the trained model uses local measurements from the target bus to predict voltage trajectory intervals in real-time.

The real-time prediction of voltage trajectory intervals presents several significant challenges: 
\begin{enumerate}
    \item \textit{Infinite-Dimensional Problem}: Voltage trajectories are continuous functions, making their prediction inherently an infinite-dimensional problem. Such problem can be addressed by operator regression tools where both the input and output are variable continuous functions.
    \item \textit{Heterogeneous Input Data}: The random nature of fault clearing time, i.e., $t_{cl}$ varies due to delays or sensing/acting equipment failures, resulting in highly variable input domains. This variability challenges the standardization of the problem and requires models that can adapt to different domains.
    \item \textit{Predictive Interval Estimation}: The method should be capable of estimating predictive intervals for practical uncertainty quantification. This ensures that the model provides reliable and informative predictions for real-time decision-making tools.

    \item \textit{Low Data Regime}: There is often a limited dataset available for predicting post-fault voltage at a target bus. This scarcity of data can hinder the training process of the models, necessitating a method that effectively enables the model to learn from other sources, such as neighboring buses.
    
    \item \textit{Privacy Concerns}: When using voltage information from neighbor buses to mitigate the low data regime, it is crucial to protect the privacy of each bus's data. Ensuring data privacy while still using distributed information for model training is a significant challenge that must be addressed.
    \item \textit{Robustness Requirements}: Reliable predictions are crucial for maintaining the stability of the system. Post-fault voltage prediction methods must deliver robust predictions under various conditions. In addition, providing coverage guarantees for predictive intervals is highly valuable, yet challenging. Achieving this combination of robustness and coverage is essential for effective power system management, although it is difficult to achieve.
\end{enumerate}
\section{Methodology} \label{sec:methodology}
In this section, we present our comprehensive approach to addressing the challenges of predicting post-fault voltage trajectories in power systems. To handle the infinite-dimensional nature of voltage prediction, our methodology leverages deep operator learning-based models. Considering the dynamics of the problem, we propose a modified DeepONet-based model, adapting the vanilla DeepONet to make it suitable for predicting voltage trajectory intervals. This modification allows the model to manage heterogeneous input data and ensures effective adaptation to varying input domains. The proposed model integrates uncertainty quantification capabilities by estimating predictive intervals and providing reliable and informative predictions. To address low data regimes, we utilize a pre-training and fine-tuning process, which can maintain privacy by enabling the model to learn from distributed data sources while maintaining data confidentiality. Finally, we employ conformal prediction techniques to ensure robust and reliable predictions by providing coverage guarantees. The following sections provide detailed descriptions of each of these techniques and strategies.

\subsection{Neural Operators}
To achieve our goal of reliably predicting post-fault voltage trajectories, we need to develop a model that can handle the infinite-dimensional nature of this problem. To this end, we employ neural operators, in particular, deep operator learning techniques to handle this complexity effectively.
Accordingly, in the next sections, we first formalize the problem as an operator regression task. Next, we outline the vanilla Deep operator network structure and then provide a step-by-step explanation of how we modify this structure to develop our model, which is designed for predicting post-fault voltage trajectory prediction intervals.

\subsubsection{Operator Regression Formulation}

We aim to approximate an operator \( \mathcal{G} \) that maps an input function \( u(t) \in \mathcal{U} \) to its corresponding post-fault trajectory function \( v(t) \in \mathcal{V} \). Formally, the operator \( \mathcal{G} \) is defined by the following mapping:

\begin{equation}
\mathcal{G} : \mathcal{U} \rightarrow \mathcal{V}
\end{equation}

where \( \mathcal{U} \) represents the vector space of the input functions, and \( \mathcal{V} \) represents the vector space of the output functions, specifically the post-fault voltage trajectories.

In the dynamics of the power grid, as described in Section \ref{sec:Problem_Definition}, the dynamic states \( x \) and the algebraic states \( z \) are not fully observable. Consequently, we used the system's observable data to define the input function space in the mapping \(\mathcal{G}\). Specifically, we define the observable portion of voltage trajectories as the input function \(u \in \mathcal{U}\) in this mapping. This allows us to utilize the voltage's past behavior to predict its future behavior, particularly its corresponding post-fault trajectory. Conceptually, this approach aligns with the Mori-Zwanzig formalism which addresses the dynamics of complex systems by focusing on observable quantities \citep{chorin2000optimal}. In this context, \( \mathcal{G} \) uses the observed data to infer the dynamic response of the system to the fault, without direct access to the internal states.

To define the input function \(u \in \mathcal{U}\) and the target solution \(v \in \mathcal{V}\), we split the voltage trajectory, denoted as \(\mathcal{R}(t)\), into two segments. We assume that the first segment is observable and consider it as the input function of the mapping \(\mathcal{G}\), while the second segment, which is unobservable, is our target solution to predict.

Since the pre-fault and fault-on conditions of the system influence the transient response in the post-fault stage, it is intuitive that the time domain of the input function, denoted as \(\mathcal{T}_u\), should encompass both of these stages of the voltage's trajectory. Additionally, to accurately map and understand the system response to a fault, we assume that a small portion of the post-fault stage is observable, referred to as the observable post-fault duration, denoted as \(\Delta t_{\text{obs}}\).

Consequently, we define the input function time domain \(\mathcal{T}_u\) as follows:
\begin{equation}
\mathcal{T}_u = [t_0, t_f) \cup [t_f, t_{cl}) \cup [t_{cl}, t_{cl} +  \Delta t_{\text{obs}}]
\label{input_domain}
\end{equation}
where \(\mathcal{T}_u\) includes the pre-fault stage \([t_0, t_f)\), the fault-on stage \([t_f, t_{cl})\), and the observable part of the post-fault trajectory \([t_{cl}, t_{cl} + \Delta t_{\text{obs}}]\). Notably, \(\Delta t_{\text{obs}}\) is an adjustable hyperparameter that directly affects the performance of the mapping by the operator \(\mathcal{G}\). Carefully selecting \(\Delta t_{\text{obs}}\) is crucial for optimizing the model's prediction performance and usability. While a smaller \(\Delta t_{\text{obs}}\) is advantageous for practical use, as the model will require less information as input; however, making it too small may negatively impact the model's performance.

\begin{figure}[!h]
  \centering
  \includegraphics[width=0.75\textwidth]{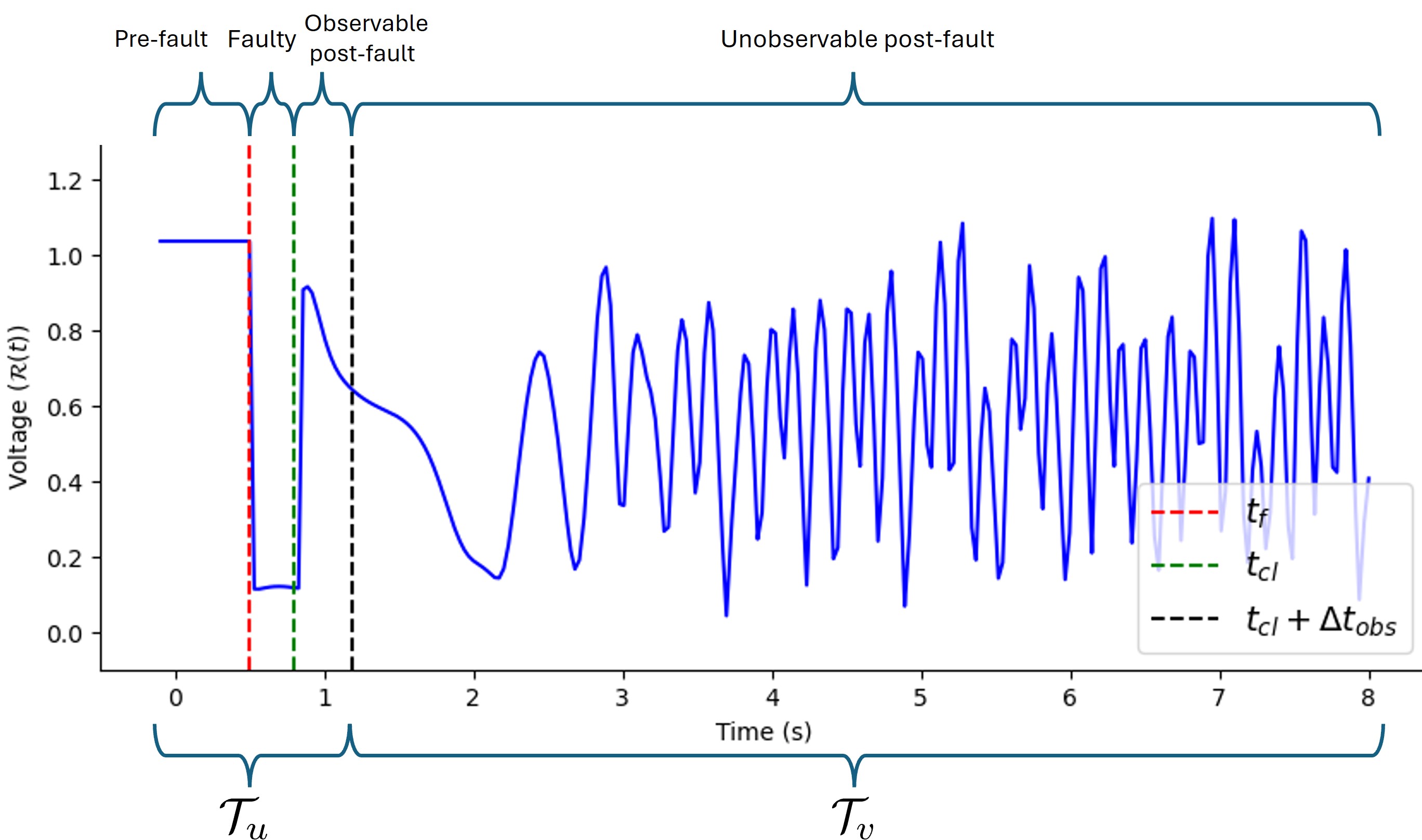}
  \caption{Temporal domains \(\mathcal{T}_u\) and \(\mathcal{T}_v\) for a sample voltage trajectory. The input function \(u\) is defined over \(\mathcal{T}_u\), and the output function \(v\) is defined over \(\mathcal{T}_v\). \(\Delta t_{\text{obs}}\) is an adjustable hyperparameter that determines the duration of observable part of the post-fault stage, directly influencing the performance of the mapping \(\mathcal{G}\).}
  \label{Picture1}
\end{figure}

The output function time domain, denoted as \(\mathcal{T}_v\), is defined as the complement of \(\mathcal{T}_u\), representing the time domain of the unobservable part of the post-fault trajectory:
\begin{equation}
\mathcal{T}_v = (t_{cl} + \Delta t_{\text{obs}}, T]
\end{equation}
where \(T\) is the maximum time of prediction. Based on this definition, for a sample trajectory \(\mathcal{R}(t)\), the input function is \(u = {\mathcal{R}}[\mathcal{T}_u]\) and the output function is \(v = {\mathcal{R}}[\mathcal{T}_v]\). Figure~\ref{Picture1} illustrates the temporal domains \(\mathcal{T}_u\) and \(\mathcal{T}_v\) for a sample voltage trajectory.

\subsubsection{Deep Operator Network} 
\begin{figure}[!h]
  \centering
  \includegraphics[width=0.7\textwidth]{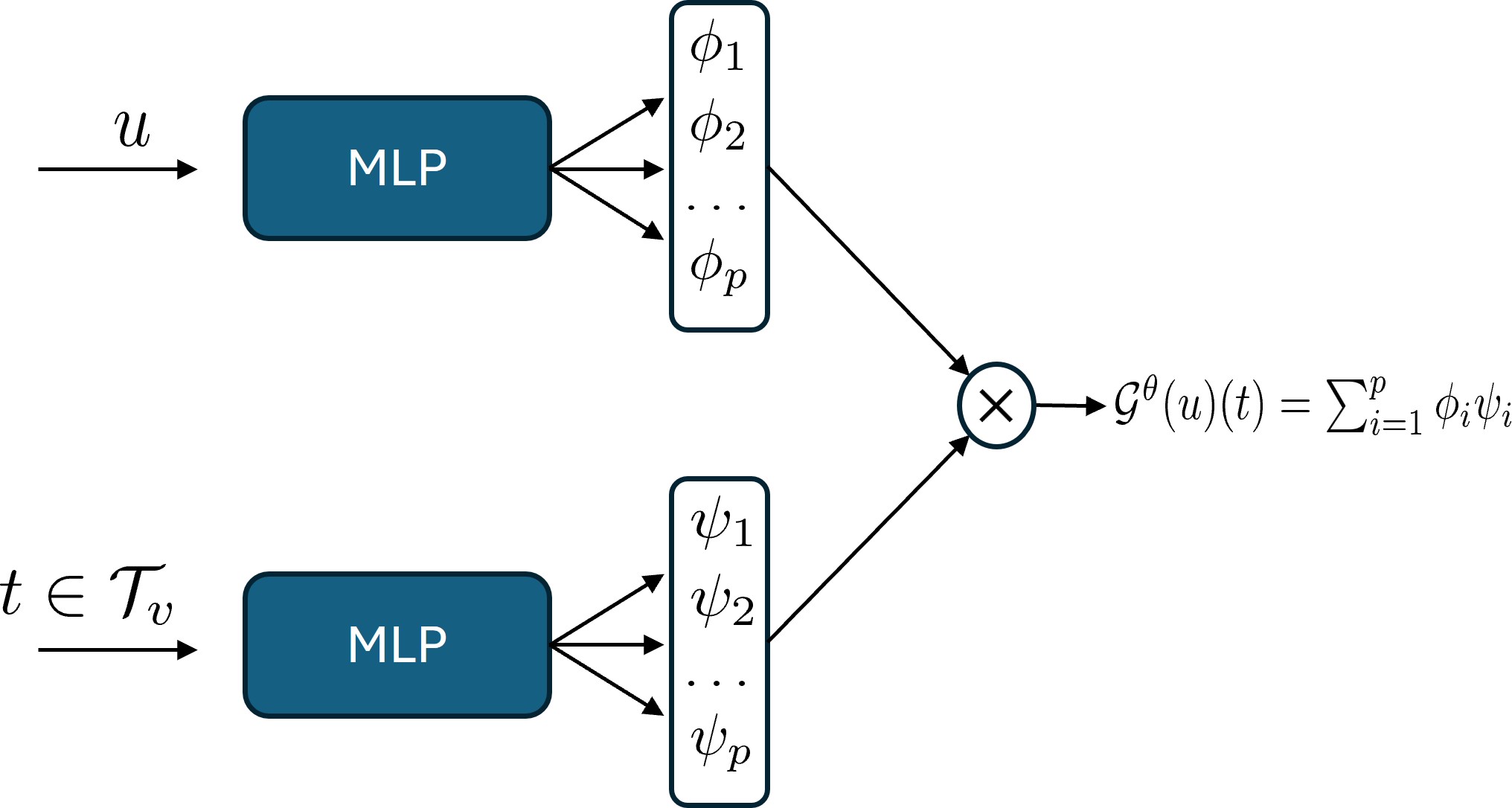}
\caption{A schematic representation of the DeepONet architecture. The figure illustrates the two main components: the branch network, which encodes the input function \( u \in \mathcal{U} \), and the trunk network, which processes the evaluation points \( t \in \mathcal{T}_v \). The outputs of these networks are merged through an inner product to approximate the target operator \(\mathcal{G} \approx \mathcal{G}^{\theta}.\)}
  \label{fig_vanilla_DeepONet}
\end{figure}

 DeepONet is an advanced neural network architecture designed to approximate operators by extending the universal approximation theorem to functions \citep{1995Chen1,2021Lulu1}. DeepONet approximates the target operator \(\mathcal{G}\) using two multilayer perceptron (MLP) sub-networks: the branch network, which encodes the input function, and the trunk network, which processes the evaluation points at the output domain. Subsequently, the outputs of these networks are combined through a linear product to approximate the operator. Given \( u \in \mathcal{U} \) as the input function and \( t \in \mathcal{T}_v \) as the evaluation point in the time domain of output, we approximate the unobserved post-fault trajectory $v$ using the following linear representation:
\begin{equation}
v(u,t) \approx \mathcal{G}^{\theta} (u)(t) = \sum_{i=1}^p \phi_i(u) \psi_i(t)
\label{equ:DeepONet}
\end{equation}
Here, \( \phi_i \) and \( \psi_i \) represent the basis functions generated by the branch and trunk networks, respectively, while \( p \) denotes the number of basis functions. Figure~\ref{fig_vanilla_DeepONet} illustrates the components of DeepONet and their integration.
However, the current configuration of the DeepONet architecture utilizes MLP networks, which may not be suitable for processing input functions that contain temporal sequences. Moreover, the post-fault voltage trajectories contain fluctuation patterns that are highly challenging for an MLP to capture accurately. 

\subsubsection{Attention-Fourier Deep Operator Network (AF-DeepONet)}

\begin{figure}[!h]
  \centering
  \includegraphics[width=0.95\textwidth]{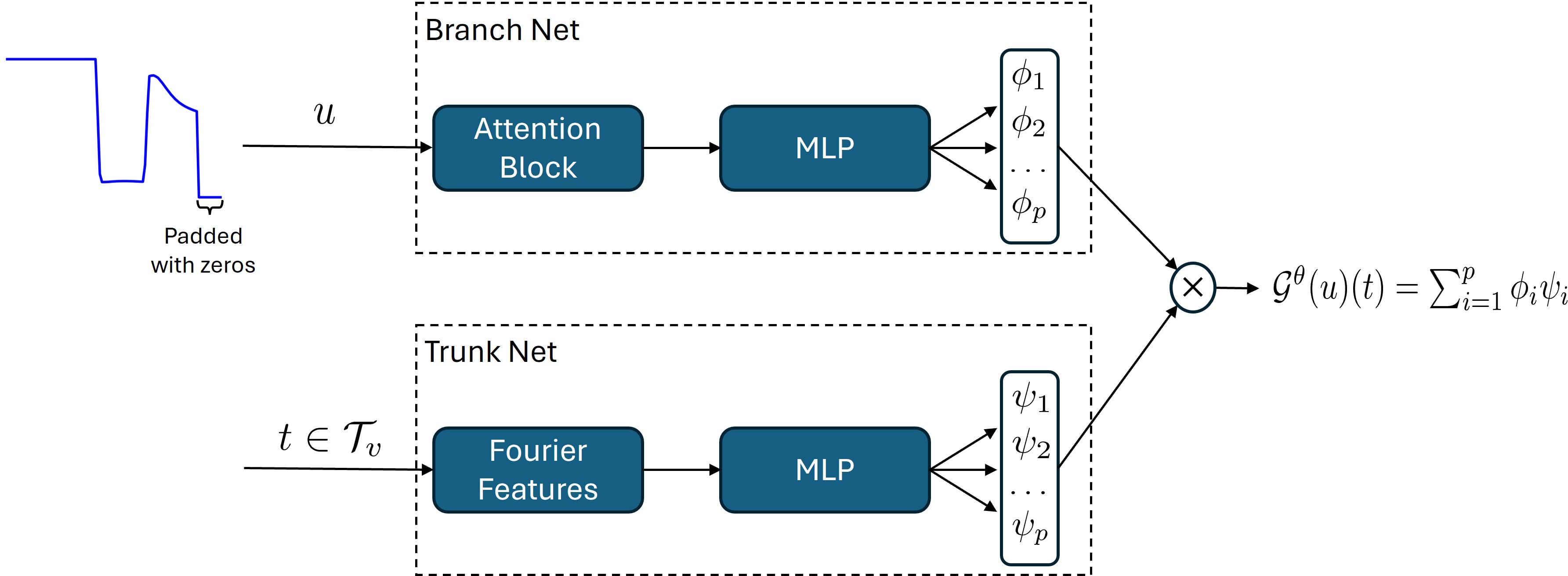}
  \caption{Schematic representation of the Attention-Fourier Deep Operator Network (AF-DeepONet).}
  \label{fig_enhanced_DeepONet}
\end{figure}

To address the limitations of DeepONet in predicting post-fault voltage trajectories, we introduce AF-DeepONet, an enhanced version of DeepONet that more effectively processes input functions and comprehends fluctuations in voltage trajectories. Figure~\ref{fig_enhanced_DeepONet} illustrates the architecture of AF-DeepONet, which consists of two primary components: the Attention-Based Branch Network and the Fourier-Featured Trunk Network.

\textit{Attention-Based Branch Network:} The branch network integrates a self-attention mechanism followed by a multi-layer perceptron (MLP) to process the input function \( u \). In particular, given a projected padded input~$\hat{u} \in \mathbb{R}^{T \times d}$, the self-attention uses trainable linear projections to compute the queries, keys, and values, i.e., $Q,K,V \in \mathbb{R}^{T \times d}$. Then, we compute the self-attention output as follows:
$$O = \text{softmax} \left( QK^\top\right) \cdot V \in \mathbb{R}^{T \times d}$$.
By employing self-attention, it selectively focuses on the most relevant parts of the input function, thereby enhancing its ability to capture dependencies and temporal dynamics. This selective weighting of input function components ensures an effective approximation of the target operator. The subsequent MLP refines these features to produce the basis functions (\( \phi_i \)).

\textit{Fourier-Featured Trunk Network:} The trunk network encodes the output location \( t \in \mathcal{T}_v \) using a Fourier features network as proposed by \citep{tancik2020fourier}. This network helps DeepONet to capture fluctuations and oscillatory behaviors, which is particularly useful for the post-fault voltage stage, which often contains high-frequency features \citep{fluids8120323}. The Fourier features network employs a random Fourier mapping:
\begin{equation}
\gamma(x) = [\sin(Bx), \cos(Bx)]
\end{equation}
where \( B \in \mathbb{R}^{m \times d} \) is a matrix with entries sampled from a Gaussian distribution \( N(0, \sigma^2) \). The encoded output location is then processed by an MLP to produce the basis functions \( \psi_i \).

Ultimately, the AF-DeepONet approximates the target operator \( \mathcal{G} \) using the same linear trainable representation as described in Equation \ref{equ:DeepONet}.

\textit{Handling Variable Input Lengths:} Given that the fault clearing time, i.e., $t_{cl}$ is a random variable, the length of \(\mathcal{T}_u\) is also random (see equation \ref{input_domain}), with the pre-fault stage having a variable duration across different voltage trajectories. This variability conflicts with the network architecture, which requires a fixed domain for its input. To address this, we employ a padding technique. We consider a maximum possible input function time \(T_{max}\), based on the maximum time for fault clearance. In this approach, we pad each voltage trajectory data to this maximum length with zero-masked signals, where the actual input function occupies the initial part and the remainder is filled with zeros, i.e., \(\mathcal{T}_u \subseteq [t_0, T_{max}]\). This approach ensures that the network can handle variable input lengths without compromising its structure.

\subsubsection{Quantile Attention-Fourier Deep Operator Network (QAF-DeepONet)}\label{subsec:QAF-DeepONet}

Power systems require robust predictive intervals to ensure stability and reliability under dynamic conditions. These intervals aid operators in managing instability and improving decision-making under uncertainty. This necessity drives our approach, underscoring the critical role of estimating uncertainty for stable and reliable forecasting. Consequently, it is essential to modify the AF-DeepONet framework to enable practical uncertainty quantification. To this end, we introduce the Quantile Attention-Fourier Deep Operator Network (QAF-DeepONet), which enhances AF-DeepONet to empirically approximate uncertainty in post-fault trajectory predictions.

Quantiles are particularly effective for uncertainty quantification because they do not assume any specific distribution, allowing them to handle distribution-free uncertainty. This inherent flexibility makes quantiles robust against skewness, heavy tails, and outliers, ensuring their applicability across various data types and scenarios. The conditional quantile function, \( q_\tau \), defines the \(\tau\)-th percentile in the distribution of \( Y \), conditional on \( X = x \). Mathematically, this function is expressed as:
\begin{equation}
q_\tau(x) = \inf \{ y \in \mathbb{R} : \mathbb{P}(Y \leq y \mid X = x)  \geq \tau \}
\end{equation}
In essence, given \(X = x\), \( q_\tau \) provides the value below which a fraction \(\tau\) of the conditional distribution of \(Y\) lies.

Building on the AF-DeepONet framework, the QAF-DeepONet is specifically designed to approximate the conditional quantiles of the target solution. For a specified level of miscoverage \(\alpha \in (0, 1)\), the QAF-DeepONet estimates the \(\alpha/2\)-th and \((1-\alpha/2)\)-th conditional quantiles, denoted as \(q_{\alpha/2}\) and \(q_{1-\alpha/2}\), respectively.

\begin{figure*}[!h]
  \centering
  \includegraphics[width=0.9\textwidth]{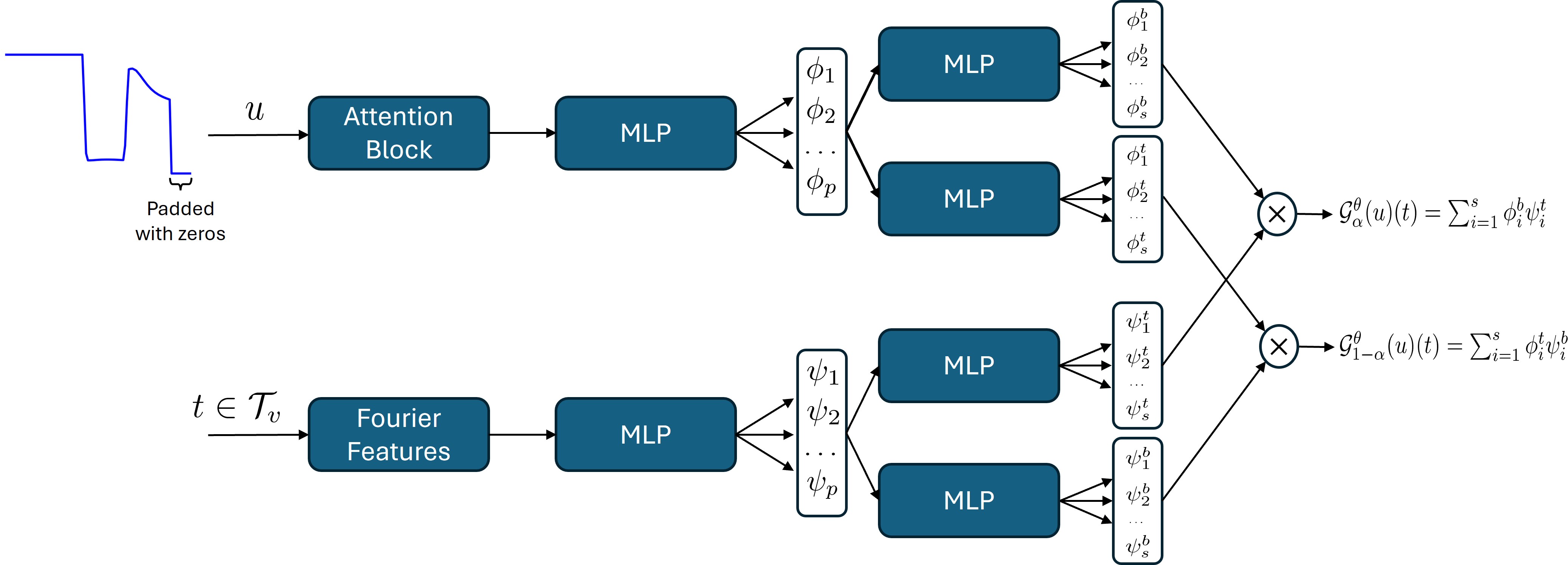}
  \caption{Schematic representation of the Quantile Attention-Fourier Deep Operator Network (QAF-DeepONet).}
  \label{Quantile_Atten_FF_DON}
\end{figure*}

Figure~\ref{Quantile_Atten_FF_DON} illustrates the QAF-DeepONet architecture. This architecture introduces additional branches and trunks to AF-DeepONet architecture specifically designed to approximate the \( q_{\alpha/2} \) and \( q_{1-\alpha/2} \) quantiles. In this configuration, the basis functions \( \phi \) and \( \psi \), generated by the original AF-DeepONet, flow into these new shared branches and trunks. These new sub-networks then process the existing basis functions to construct new ones, capable of estimating the specified quantiles through their following inner products:
\begin{equation}
q_{\alpha/2}(u,t) \approx \mathcal{G}^{\theta}_\alpha (u)(t) = \sum_{i=1}^{s} \phi^b_i \psi^t_i
\end{equation}
\begin{equation}
q_{1-\alpha/2}(u,t) \approx \mathcal{G}^{\theta}_{1-\alpha/2} (u)(t) = \sum_{i=1}^{s} \phi^t_i \psi^b_i
\end{equation}
Here, \(\phi^b_i\) and \(\psi^t_i\) serve as the basis functions for estimating \( q_{\alpha/2} \), while \(\phi^t_i\) and \(\psi^b_i\) are used for estimating \( q_{1-\alpha/2} \). The number of these basis functions is denoted by \( s \).

\textit{Dataset Generation:} To generate input and output functions, we first choose the observable post-fault duration \(\Delta t_{\text{obs}}\). For each voltage trajectory \( \mathcal{R}_j \), considering the clearing time \( t_{cl} \) and chosen \(\Delta t_{\text{obs}}\), we split \( \mathcal{R}_j \) into the observable trajectory \( u_j = \mathcal{R}_j[\mathcal{T}_u] \) and the unobservable trajectory \( v_j = \mathcal{R}_j[\mathcal{T}_v] \). We then pad \( u_j \) with zeros to a predefined length to create the input function. For convenience in this paper, we refer to this padded version of \( u_j \) same as \( u_j \). The output function \( v_j \) represents the unobserved post-fault trajectory.

Furthermore, for both the training and the inference phases in QAF-DeepONet, it is essential to structure the datasets to be compatible with the DeepONet configurations. To achieve this, the continuous function \( u_j \) is discretized at the \( m \) sensor locations, that is, \(u_j \in \mathbb{R}^m \), which serves as input to the branch net. Moreover, for each \( u_j \), we randomly select \( n_{loc} \) number of output locations \( t \in \mathcal{T}_v \) and sample the output function at these points, that is, \( \{G^{(i)}\}_{i=1}^{n_{loc}} = \{v_j(t^{(i)})\}_{i=1}^{n_{loc}} \). This approach is consistently applied across all voltage trajectories to assemble a dataset of triplets, denoted as \( \mathcal{D} = \{u^{(i)}, t^{(i)}, G^{(i)}\}_{i=1}^N \).

\textit{Loss Function:} To optimize the QAF-DeepONet, we minimize the following loss function across a dataset of \( N \) triplets, denoted as \( \mathcal{D} = \{u^{(i)}, t^{(i)}, G^{(i)}\}_{i=1}^N \):

\begin{equation}
\mathcal{L}(\theta, \mathcal{D}) = \frac{1}{N} \sum_{i=1}^N \rho_\tau\left(G^{(i)}, \mathcal{G}_\tau\left(u^{(i)}\right)(t^{(i)})\right),
\end{equation}
where \( \rho_\tau \) represents the pinball loss function for quantiles \( \tau \in \{\alpha/2, 1-\alpha/2\} \). The function is specifically defined as follows:
\begin{equation}
\rho_\tau(y, \hat{y}) = \begin{cases} 
\tau (y - \hat{y}) & \text{if } y - \hat{y} > 0, \\
(1 - \tau) (\hat{y} - y) & \text{otherwise}.
\end{cases}
\label{equ:ro}
\end{equation}
This loss function allows the evaluation of model performance based on quantile-specific discrepancies between the predicted intervals and actual values.

After training, the QAF-DeepONet estimates predictive intervals for any new test sample as follows:
\begin{equation}
  \widehat{CL}(\hat{u}_\text{test},x_\text{test}) = \left[\mathcal{G}^{\theta}_{\alpha/2}\left(u_\text{test}\right)\left(t_\text{test}\right), \mathcal{G}^{\theta}_{1-\alpha/2} \left(u_\text{test}\right)\left(t_\text{test}\right) \right],
  \label{equ:CL}
\end{equation}
where \(\mathcal{G}^{\theta}_{\alpha/2}\left(u_\text{test}\right)\left(t_\text{test}\right)\) and \(\mathcal{G}^{\theta}_{1-\alpha/2}\left(u_\text{test}\right)\left(t_\text{test}\right)\) denote the predicted lower and upper quantiles, respectively, for the interval with miscoverage level \(\alpha\).

\subsection{Training Strategy}

In power systems, faults are rare events, which makes it impractical to obtain extensive fault data for various scenarios. Furthermore, data from the power system are often private and not easily accessible, leading to limited data availability. It is particularly challenging to obtain voltage trajectory data for a specific bus. However, data-driven approaches, such as deep operator learning, often require large datasets for high performance.

To address this issue, our methodology utilizes a pre-training and fine-tuning strategy, making maximum use of the available data in the system. In this strategy, we first pre-train the model on a broader dataset from neighboring buses to learn the underlying voltage dynamics. We then fine-tune the model using data from the target bus, aiming for reliable post-fault trajectory predictions. Figure~\ref{fig:39_bus_system} shows a schematic of a power system, highlighting the target bus and its neighboring buses.

\subsubsection{Pre-Training Model}

The pre-training phase is crucial in our methodology, providing the foundation for developing a robust and accurate model. We utilize data from neighboring buses of the target bus to pre-train the model. However, due to privacy concerns, costs, and logistical constraints, direct data sharing is not feasible. To overcome this, we employ federated learning, enabling collaborative training of local models to obtain a \textit{merged} model without direct data sharing.

\textit{Federated Deep Operator Network Learning:} Federated learning effectively addresses privacy and related issues by decentralizing the training process. In this strategy, each bus's model is trained locally, sharing only its parameters \(\theta\) with other buses, thus preserving data privacy. To implement this strategy, we have adapted the approach outlined in \citep{a15090325} and improved in~\citep{zhang2024d2no}.

Consider a target bus with \(\mathcal{N}_c\) neighboring buses. Let \(\theta_k^c\) represent the parameters of the model for the \(c\)-th bus in iteration \(k\). The parameters \(\gamma_k^c\) are the updated values derived from \(\theta_k^c\) through a local Adam optimization step at the same iteration. With \(K\) denoting the number of local round updates, the model's parameters are updated using the following procedure:
\begin{equation}
\theta_{k+1}^c = \begin{cases}
\gamma_{k+1}^c & \text{if } (k + 1)\mod K \neq 0 \\
\sum_{c=1}^{\mathcal{N}_c} \frac{1}{\mathcal{N}_c} \gamma_{k+1}^c & \text{if } (k + 1)\mod K = 0
\end{cases}
\end{equation}
In this procedure, bus \(c\) independently updates its parameters using local Adam optimization. For most iterations, the updated parameters \(\gamma_{k+1}^c\) are directly adopted as the new parameters \(\theta_{k+1}^c\) for the local model of the bus \(c\). However, after every \(K\) local update, the parameters are averaged across all neighboring buses to form a new global parameter set. Then, this averaged model becomes the updated local model for all buses. After training, the averaged model serves as the pre-trained model for the target bus, benefiting from the collective knowledge of neighboring buses and enhancing its robustness and generalization.

\textit{Pre-Training Dataset:} The federated learning dataset comprises voltage trajectories from the neighboring buses of the target bus. Each neighboring bus dataset contains its own voltage trajectory data, used locally to train its corresponding local model. The pre-training dataset for \(\mathcal{N}_c\) neighboring buses is represented as \(\mathcal{D}_\mathcal{P} = \{\mathcal{D}_c\}_{c=1}^{{\mathcal{N}_c}}\), where \(\mathcal{D}_c = \{u^{(i)}, t^{(i)}, G^{(i)}\}_{i=1}^{n_c}\) denotes the dataset for bus \(c\), with \(n_c\) being the number of training triples for that bus. The size of the pre-training dataset is  \(n_p = \sum_{\mathcal{C}=1}^{\mathcal{N}_c} n_c\).

\subsubsection{Finetuning the Pre-Trained Model} 
After pre-training the model, we proceed to fine-tune it using data from the target bus. Since the pre-trained model has already captured the underlying dynamics of the voltage trajectories, fine-tuning allows it to adapt to the unique attributes of the target bus using a relatively small dataset, effectively overcoming the challenge of limited data availability. Through this process, the model learns the specific dynamics and operational characteristics of the target bus, thereby enhancing its precision and reliability.

\textit{Fine-Tuning Dataset:} The fine-tuning dataset consists of voltage trajectories from the target bus. We transform each of these trajectories into QAF-DeepONet-compatible formats and assemble the fine-tuning dataset \(\mathcal{D}_\mathcal{F} = \{u^{(i)}, t^{(i)}, G^{(i)}\}_{i=1}^{n_f}\), where \(n_f\) is the number of training triples for the fine-tuning dataset.

Let \(\bar{\theta}\) represent the parameters of the pre-trained QAF-DeepONet. To fine-tune the model for the target bus, a local update is performed on \(\bar{\theta}\) using the target bus dataset \(\mathcal{D}_\mathcal{F}\), applying local Adam optimization to update the parameters to the fine-tuned values, denoted as \(\theta^f\).

It is important to note that the pre-trained model contains extensive information about the local voltage dynamics; therefore, the fine-tuning process must be proceed carefully to prevent catastrophic forgetting. To this end, an early stopping technique is implemented, which helps the model adapt to the dynamics of the target bus without forgetting the core features learned during the pre-training process.

\subsection{Conformal Prediction with QAF-DeepONet}

The predictive interval estimated in Equation \ref{equ:CL} often fails to meet the desired coverage probability criterion. To address this issue, our proposed methodology integrates the Split Conformal Quantile-DeepONet Prediction methods introduced in \citep{moya2024conformalized}. This approach ensures that the prediction intervals are calibrated to achieve the pre-specified coverage probability, thereby improving the reliability and robustness of the predictions.

This is particularly important in power systems, where accurate and reliable post-fault trajectory predictions are crucial for system stability and operational planning. Guaranteeing coverage probability of the predicted intervals enhances the assessment of risks and uncertainties associated with fault events, leading to more informed decision-making and proactive measures. This strengthens the overall resilience of the power system, enabling operators to anticipate and mitigate potential issues more effectively.

To ensure the coverage guarantee, the conformal approach employs a calibration dataset \(\mathcal{D}_{cal} = \{u^{(i)}, t^{(i)}, G^{(i)}\}_{i=1}^{n_{cal}}\) for calibrating model predictions. Utilizing a trained QAF-DeepONet, which approximate quantile operators \(\mathcal{G}^{\theta}_{\alpha/2}\) and \(\mathcal{G}^{\theta}_{1-\alpha/2}\) for a specified miscoverage level \(\alpha \in (0,1)\), we first define the score function for calibration data triplets \((u^{(i)}, t^{(i)}, G^{(i)})\) as follows:
\begin{equation}
s(u, t, G) = \max \left\{\mathcal{G}^{\theta}_{\alpha/2}(u)(t) - G, G - \mathcal{G}^{\theta}_{1-\alpha/2}(u)(t)\right\}
\end{equation}
This score function measures the error incurred by QAF-DeepONet when the operator target $G$ falls outside the estimated quantiles. Using this function, we compute the calibration scores for the entire calibration dataset $\{u^{(i)}, t^{(i)}, G^{(i)}\}_{i=1}^{n}$: 
$s_1 = s(u^{(1)}, t^{(1)}, G^{(1)}), \dots, s_n = s(u^{(n)}, t^{(n)}, G^{(n)})$.

Next, we calculate the calibration factor, \(\hat{q}\), as the \(\frac{\lceil (n+1) (1-\alpha) \rceil}{n}\)-th quantile of the calibration scores \(s_1, \dots, s_n\). Finally, for any new test data \((u_\text{test}, t_\text{test})\), we modify the original estimated predictive interval via Equation \ref{equ:CL} to the calibrated predictive interval:
\begin{equation}
\widehat{CL}_{\text{c}}(u_\text{test}, t_\text{test}) = \left[\mathcal{G}^{\theta}_{\alpha/2}(u_\text{test})(t_\text{test}) - \hat{q}, \mathcal{G}^{\theta}_{1-\alpha/2}(u_\text{test})(t_\text{test}) + \hat{q}\right].
\end{equation}
This predicted interval is adaptive, distribution-free, and guarantees coverage probability of predicted intervals, ensuring reliable performance. The summary of the described approach is detailed in Algorithm \ref{alg:1}.

\begin{algorithm}[!h]
\DontPrintSemicolon
\LinesNumberedHidden
\caption{Split Conformal QAF-DeepONet Prediction}
\label{alg:CQR-DeepONet}
\KwIn{Trained QAF-DeepONet model, Calibration dataset \(\mathcal{D}_{\text{cal}} = \{u^{(i)}, t^{(i)}, G^{(i)}\}_{i=1}^{n_{\text{cal}}}\), Miscoverage level \(\alpha\)}
\KwOut{Calibrated prediction interval for test data \((u_\text{test}, t_\text{test})\)}
\BlankLine
\textbf{Step 1: Calculate calibration scores} \;
\Indp for {$i = 1, \dots, n$} \textbf{do}\\
\Indp{
    \small{
    $s_i = \max \left\{\mathcal{G}^{\theta}_{\alpha/2}(u^{(i)}, t^{(i)}) - G^{(i)}, G^{(i)} - \mathcal{G}^{\theta}_{1-\alpha/2}(u^{(i)}, t^{(i)})\right\}$ \;
    }
}
\BlankLine
\Indm \Indm \textbf{Step 2: Calculate calibration factor} \;
\Indp Sort $s_1, s_2, \dots, s_n$ in ascending order, then calculate the calibration quantile:\;
\(\hat{q} \leftarrow \frac{\lceil (n+1) (1-\alpha) \rceil}{n}\)-th quantile of the calibration scores \(s_1, \dots, s_n\)\;
\Indm
\BlankLine
\textbf{Step 3: Calibrate prediction interval} \;
\Indp for {new test data \((u_\text{test}, t_\text{test})\)} compute:\\
\small{
    \(\widehat{CL}_{\text{c}}(u_\text{test}, t_\text{test}) = \left[\mathcal{G}^{\theta}_{\alpha/2}(u_\text{test}, t_\text{test}) - \hat{q}, \mathcal{G}^{\theta}_{1-\alpha/2}(u_\text{test}, t_\text{test}) + \hat{q}\right]\)\;
}
\BlankLine
\Indm
\Return \(\widehat{CL}_{\text{c}}(u_\text{test}, t_\text{test})\)
\label{alg:1}
\end{algorithm}
\section{Numerical Experiments} \label{sec:Numerical_Experiments}
In this section, we assess the performance of our proposed methodology in estimating predictive intervals for post-fault trajectories across various scenarios.

\begin{figure*}[!h]
  \centering
  \includegraphics[width=0.9\textwidth, trim=0 350 0 0, clip]{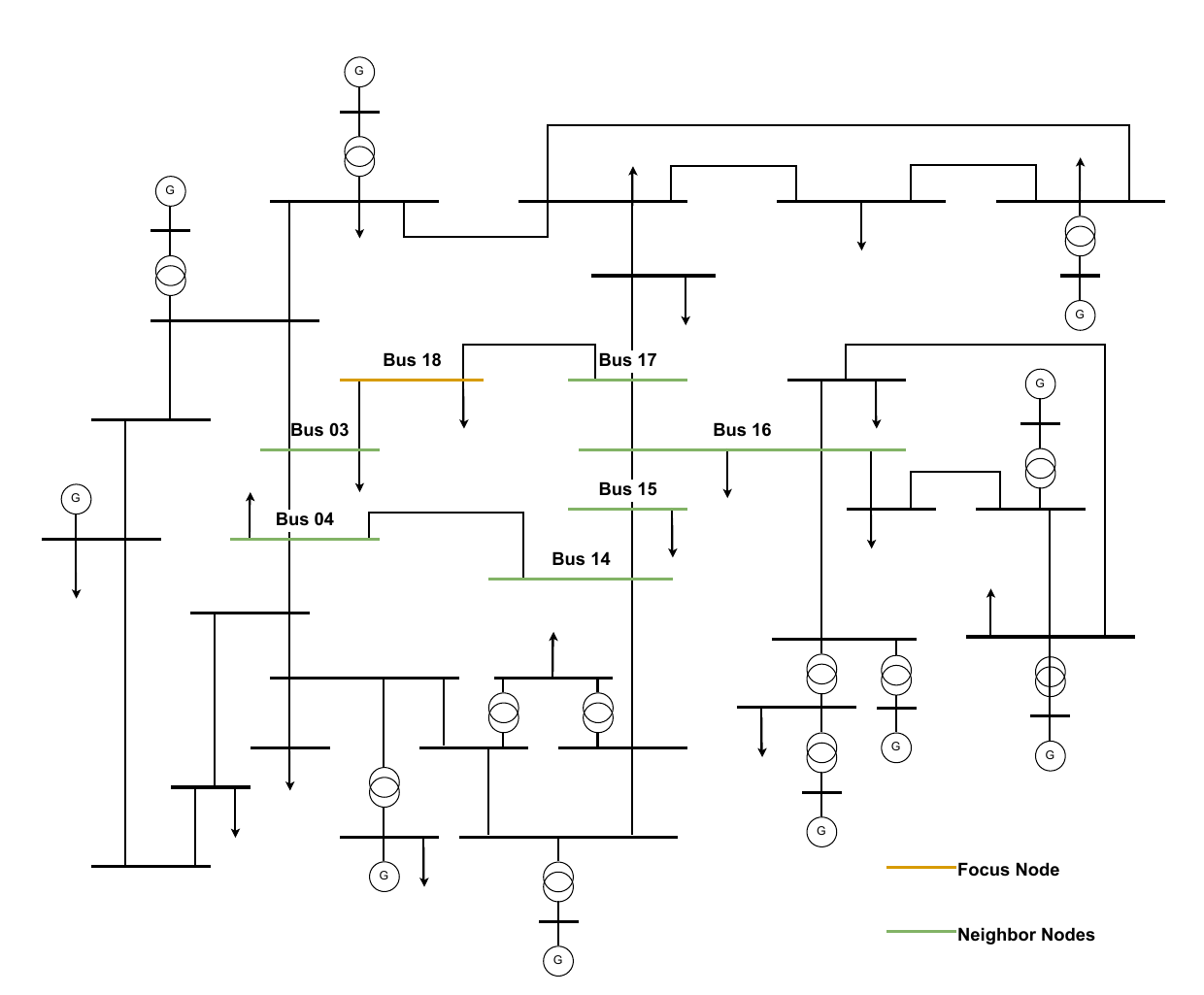}
  \caption{New England 39-bus test system.}
   \label{fig:39_bus_system}
\end{figure*}

\subsection{Data} \label{sec:Data}

We validate the proposed methodology using the New England 39-bus test system, shown in Figure~\ref{fig:39_bus_system}. Notably, we use a realistic setup where each generating unit includes detailed models for voltage and frequency controllers, specifically an automatic voltage regulator and governor. The parameters for the network, generators, and controllers are sourced from the DIgSILENT PowerFactory library.

To generate realistic datasets, voltage trajectories are generated for each bus under the following simulation conditions:

\begin{itemize}
    \item \textit{Demand variation:} Active and reactive power demands are randomly varied by $\pm 30\%$ of their nominal values.
    \item \textit{Network contingencies:} In each simulation, (i) a three-phase short circuit occurs in one of the system's transmission lines, representing an N-1 contingency, (ii) after the fault, the line is isolated following the clearing time, and (iii) the fault location is randomly chosen along the line, ranging from 1\% to 99\% of its total length.
    \item \textit{Clearing time:} The fault clearing time is uniformly sampled between 100 ms and 333 ms.
\end{itemize}

Furthermore, throughout this paper, we denote the number of voltage trajectories used for pre-training, fine-tuning, and calibration by \( N_{\mathcal{P}} \), \( N_{\mathcal{F}} \), and \( N_{cal} \), respectively.

\subsection{The Model} \label{sec:The_Model}

We employ the proposed QAF-DeepONet, configured as outlined in Section \ref{subsec:QAF-DeepONet}. The hyperparameters of the model are selected through a hyperparameter tuning process.

\subsection{Training Protocol} \label{sec:Training_protocol}

Our goal is to predict post-fault voltage trajectory intervals of a target bus in the power system. To this end, we consider the neighboring buses shown in Figure~\ref{fig:39_bus_system}. In this neighborhood, there are 7 buses; we select Bus 18 as the target bus and the remaining six buses as its neighboring ones. 

We aim to predict the $95\%$ predictive interval for the post-fault voltage trajectories of the target bus. Specifically, we designed the QAF-DeepONet to predict the $0.025$-th and $0.975$-th quantiles, which represent the lower and upper bounds of the interval. The $95\%$ predictive interval is then constructed using these two quantiles.

\subsection{Evaluation Metrics} \label{sec:Evaluation_Metrics}
To evaluate QAF-DeepONet's predicted intervals for post-fault trajectories, we use two metrics: Prediction Interval Coverage Probability (PICP) and Prediction Interval Normalized Width Average (PINAW).

\textit{Prediction Interval Coverage Probability (PICP):} PICP measures the proportion of actual values within the predicted intervals:
\begin{equation}
\text{PICP} = \frac{1}{N} \sum_{i=1}^{N} I(y_i \in [\hat{y}_{i, \text{lower}}, \hat{y}_{i, \text{upper}}]),
\end{equation}
where \(N\) is the number of data points, \(y_i\) is the actual value, \(\hat{y}_{i, \text{lower}}\) and \(\hat{y}_{i, \text{upper}}\) are the bounds of the predicted interval, and \(I\) is an indicator function. A higher PICP indicates better reliability.

\textit{Prediction Interval Normalized Width Average (PINAW):} PINAW assesses the average width of prediction intervals relative to data variability:
\begin{equation}
\text{PINAW} = \frac{1}{N} \sum_{i=1}^{N} \frac{\hat{y}_{i, \text{upper}} - \hat{y}_{i, \text{lower}}}{y_{\text{max}} - y_{\text{min}}},
\end{equation}
where \(y_{\text{max}}\) and \(y_{\text{min}}\) are the maximum and minimum observed values. Lower PINAW values suggest more precise predictive intervals.

Together, these metrics evaluate the reliability and precision of the predictive performance of the model.

It should be noted that to evaluate the models, we consider 100 test trajectories, which are unseen during the training process, and report the average PICP and PINAW values across all test trajectories in the following analyses.

\subsection{Model Evaluation}\label{Model_Evaluation}

\begin{figure}[!h]
    \centering
    \begin{subfigure}[b]{\linewidth}
        \begin{tikzpicture}
            \begin{axis}[width=\linewidth, height=4cm, xlabel={Time (s)}, ylabel={Voltage}, xmin=-0.2, xmax=8.5, ymin=-0.0, ymax=1.35, 
            xtick={0,1,...,8.5}, ytick={0,0.2,...,1.3},             
            legend style={
                at={(0.5,1.15)},
                anchor=south,
                legend columns=2, 
                column sep=1ex 
            }] 

            \addplot [smooth, black, semithick] table{Figures/FigsData/sample_figures/memory_35.txt}; 

            \addplot [smooth, blue, semithick] table{Figures/FigsData/sample_figures/G_true_35.txt}; 
            
            \addplot[name path=fine_q1,  fill=none, draw=none,forget plot] table{Figures/FigsData/sample_figures/fine_q1_35.txt};
            \addplot[name path=fine_q2, fill=none, draw=none,forget plot] table{Figures/FigsData/sample_figures/fine_q2_35.txt};
            \definecolor{mycolor}{RGB}{75,200,75}
            \addplot[fill=mycolor, fill opacity=0.5] fill between[of=fine_q1 and fine_q2];

            \addplot[name path=fine_q1_conf,  fill=none, draw=none,forget plot] table{Figures/FigsData/sample_figures/fine_q1_conf_35.txt};
            \addplot[name path=fine_q2_conf, fill=none, draw=none,forget plot] table{Figures/FigsData/sample_figures/fine_q2_conf_35.txt};
            \addplot[fill=blue, fill opacity=0.1] fill between[of=fine_q1_conf and fine_q2_conf];
             \addlegendentry[font=\scriptsize]{Observed Data}
            \addlegendentry[font=\scriptsize]{Unobserved Data }
            \addlegendentry[font=\scriptsize]{Fine-Tuned Prediction}
            \addlegendentry[font=\scriptsize]{Conformal Prediction}
            \end{axis}
        \end{tikzpicture}
        \vspace{5pt}
    \end{subfigure}  
    \begin{subfigure}[b]{\linewidth}
        \begin{tikzpicture}
            \begin{axis}[width=\linewidth, height=4cm, xlabel={Time (s)}, ylabel={Voltage}, xmin=-0.2, xmax=8.5, ymin=-0.0, ymax=1.35, 
            xtick={0,1,...,8.5}, ytick={0,0.2,...,1.3}, legend pos=south east] 

            \addplot [smooth, black, semithick] table{Figures/FigsData/sample_figures/memory_45.txt}; 

            \addplot [smooth, blue, semithick] table{Figures/FigsData/sample_figures/G_true_45.txt}; 
            
            \addplot[name path=fine_q1,  fill=none, draw=none,forget plot] table{Figures/FigsData/sample_figures/fine_q1_45.txt};
            \addplot[name path=fine_q2, fill=none, draw=none,forget plot] table{Figures/FigsData/sample_figures/fine_q2_45.txt};
            \definecolor{mycolor}{RGB}{75,200,75}
            \addplot[fill=mycolor, fill opacity=0.5] fill between[of=fine_q1 and fine_q2];

            \addplot[name path=fine_q1_conf,  fill=none, draw=none,forget plot] table{Figures/FigsData/sample_figures/fine_q1_conf_45.txt};
            \addplot[name path=fine_q2_conf, fill=none, draw=none,forget plot] table{Figures/FigsData/sample_figures/fine_q2_conf_45.txt};
            \addplot[fill=blue, fill opacity=0.1] fill between[of=fine_q1_conf and fine_q2_conf];
            \end{axis}
        \end{tikzpicture}
        \vspace{5pt}
    \end{subfigure}
    \begin{subfigure}[b]{\linewidth}
        \begin{tikzpicture}
            \begin{axis}[width=\linewidth, height=4cm, xlabel={Time (s)}, ylabel={Voltage}, xmin=-0.2, xmax=8.5, ymin=0, ymax=1.35, 
            xtick={0,1,...,8.5}, ytick={0,0.2,...,1.3}, legend pos=south west] 

            \addplot [smooth, black, semithick] table{Figures/FigsData/sample_figures/memory_34.txt}; 

            \addplot [smooth, blue, semithick] table{Figures/FigsData/sample_figures/G_true_34.txt}; 
            
            \addplot[name path=fine_q1,  fill=none, draw=none,forget plot] table{Figures/FigsData/sample_figures/fine_q1_34.txt};
            \addplot[name path=fine_q2, fill=none, draw=none,forget plot] table{Figures/FigsData/sample_figures/fine_q2_34.txt};
            \definecolor{mycolor}{RGB}{75,200,75}
            \addplot[fill=mycolor, fill opacity=0.5] fill between[of=fine_q1 and fine_q2];

            \addplot[name path=fine_q1_conf,  fill=none, draw=none,forget plot] table{Figures/FigsData/sample_figures/fine_q1_conf_34.txt};
            \addplot[name path=fine_q2_conf, fill=none, draw=none,forget plot] table{Figures/FigsData/sample_figures/fine_q2_conf_34.txt};
            \addplot[fill=blue, fill opacity=0.1] fill between[of=fine_q1_conf and fine_q2_conf];
            \end{axis}
        \end{tikzpicture}
        \vspace{5pt}
    \end{subfigure}  
    \begin{subfigure}[b]{\linewidth}
        \begin{tikzpicture}
            \begin{axis}[width=\linewidth, height=4cm, xlabel={Time (s)}, ylabel={Voltage}, xmin=-0.2, xmax=8.5, ymin=0, ymax=1.45, 
            xtick={0,1,...,8.5}, ytick={0,0.2,...,1.3}, legend pos=south west] 

            \addplot [smooth, black, semithick] table{Figures/FigsData/sample_figures/memory_83.txt}; 

            \addplot [smooth, blue, semithick] table{Figures/FigsData/sample_figures/G_true_83.txt}; 
            
            \addplot[name path=fine_q1,  fill=none, draw=none,forget plot] table{Figures/FigsData/sample_figures/fine_q1_83.txt};
            \addplot[name path=fine_q2, fill=none, draw=none,forget plot] table{Figures/FigsData/sample_figures/fine_q2_83.txt};
            \definecolor{mycolor}{RGB}{75,200,75}
            \addplot[fill=mycolor, fill opacity=0.5] fill between[of=fine_q1 and fine_q2];

            \addplot[name path=fine_q1_conf,  fill=none, draw=none,forget plot] table{Figures/FigsData/sample_figures/fine_q1_conf_83.txt};
            \addplot[name path=fine_q2_conf, fill=none, draw=none,forget plot] table{Figures/FigsData/sample_figures/fine_q2_conf_83.txt};
            \addplot[fill=blue, fill opacity=0.1] fill between[of=fine_q1_conf and fine_q2_conf];
            
            \end{axis}
        \end{tikzpicture}
    \end{subfigure}   
    \caption{Predicted $95\%$ predictive interval of post-fault trajectories using the fine-tuned model (\(N_{\mathcal{P}} = 6000\), \(N_{\mathcal{F}} = 3000\)) and corresponding conformal predictions (\(N_{\text{cal}} = 100\)) for four sample voltage trajectories with an observable post-fault duration of \(\Delta t_{obs} = 400\) ms.}
    \label{fig:Smple_figures}
\end{figure}
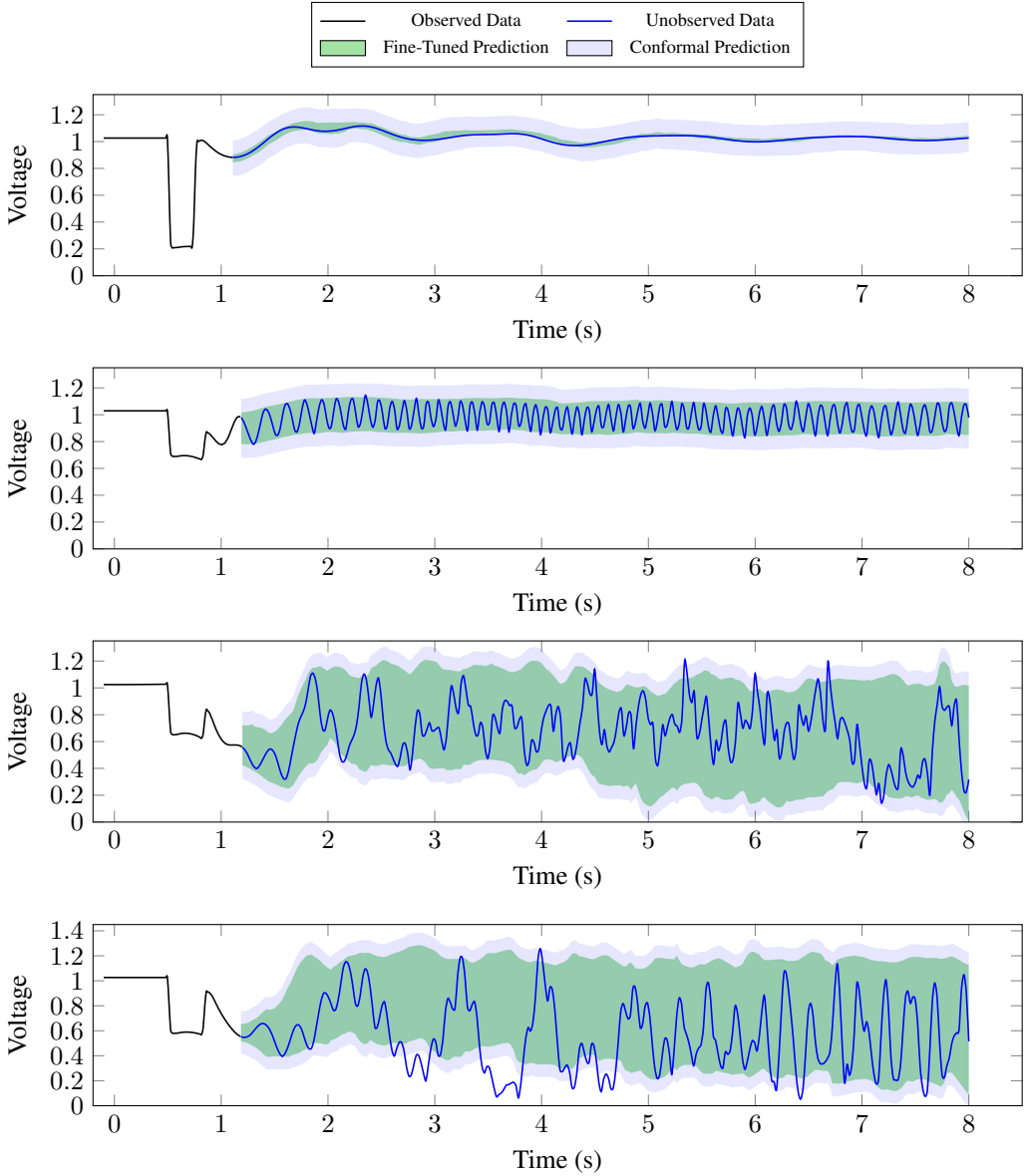

Figure~\ref{fig:Smple_figures} illustrates the prediction results of the fine-tuned model alongside the corresponding conformal predictions for four sample voltage trajectories. For these results, the model was initially pre-trained using \(N_{\mathcal{P}} = 6000\) trajectories from neighboring buses, subsequently fine-tuned with \(N_{\mathcal{F}} = 3000\) trajectories from the target bus, and finally calibrated using \(N_{\text{cal}} = 100\) trajectories from the same bus to calibrate the predictions within the conformal prediction framework. These predictions consider the scenario with a relatively short observable post-fault duration, \(\Delta t_{obs} = 400\) ms. Despite this, the fine-tuned model provides precise and reliable predictive intervals for post-fault trajectories. However, the fine-tuned prediction does not cover some parts of the true trajectory. To address this, conformal prediction adjusts the predicted interval, ensuring that most parts are covered and guaranteeing the required coverage probability, leading to more reliable predictions.

In the following sections, we analyze both pre-trained and fine-tuned models to evaluate the effectiveness of pre-training in capturing targeted mappings and the improvements achieved through fine-tuning. In addition, we evaluate the results of conformal prediction, discussing its effectiveness, advantages, and limitations.  For each model, we will perform a sensitivity analysis to understand the impact of observable post-fault duration \(\Delta t_{obs}\) and the number of trajectories used during the training process on the quality of the predicted intervals. Finally, we will compare these results to highlight key insights and discern notable differences.

\subsubsection{Evaluating the Pre-trained Model} \label{sec:Pre-trained}

We evaluated the zero-shot prediction capabilities of the pre-trained model, which means its ability to make accurate predictions without any additional fine-tuning. This is done by testing the model on data from the target bus whose dynamics was not part of the training dataset. This evaluation is crucial because it provides insight into the adaptability and potential of the model in scenarios without fine-tuning.

Table~\ref{table:zero_shot_fine_tuned} shows the average metric values for the estimated predictive intervals of the pre-trained models for post-fault trajectories on 100 test trajectories. For these predictions, we consider a scenario where the observable post-fault duration is \(\Delta t_{obs} = 500\) ms and the pre-trained model is trained on \(N_\mathcal{P} = 6000\) voltage trajectories from neighboring buses. The results show relatively high PICP values and relatively low PINAW values, highlighting the models' high performance in zero-shot prediction scenarios. This combination underscores the effectiveness of the models in providing reliable and precise predictions even without fine-tuning.

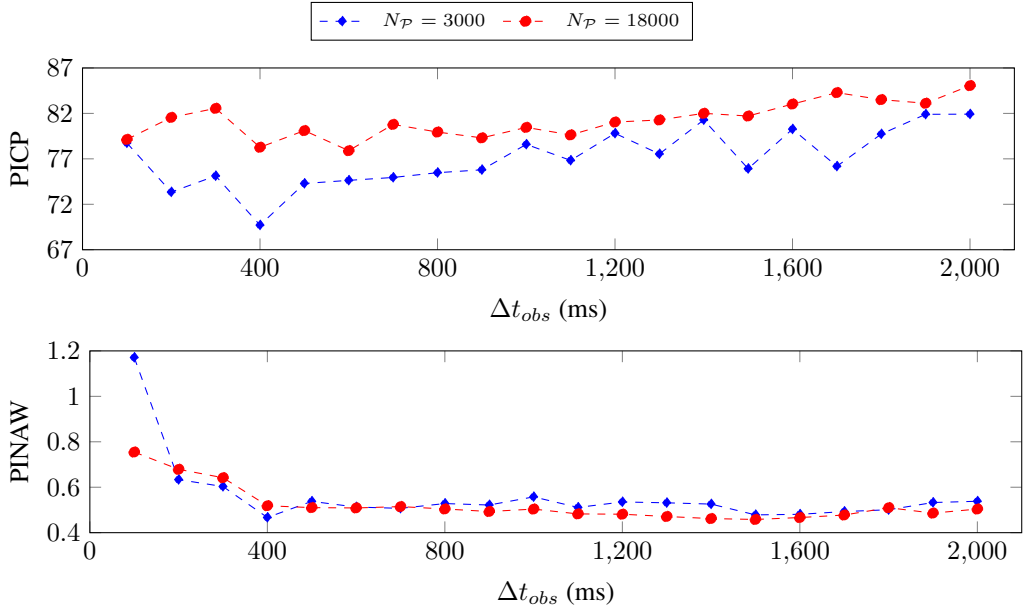
\begin{figure}[!h]
    \centering
    \begin{subfigure}[b]{\linewidth}
        \begin{tikzpicture}
            \begin{axis}[width=\linewidth, height=4cm, xlabel={$\Delta t_{obs}$ (ms)}, ylabel={PICP}, xmin=0, xmax=2100, ymin=67, ymax=87, 
            xtick={0,400,...,2000}, ytick={67,72,...,87},         
            legend style={
                at={(0.45,1.15)},
                anchor=south,
                legend columns=2, 
                column sep=1ex 
            }]    
            \addplot [dashed, blue, mark=diamond*] table{Figures/FigsData/Average_PICP500_pre.txt}; 
            \addlegendentry[font=\scriptsize]{$N_\mathcal{P} =3000$}
            \addplot [dashed, red, mark=*] table{Figures/FigsData/Average_PICP3000_pre.txt}; 
            \addlegendentry[font=\scriptsize]{$N_\mathcal{P} =18000$}
            \end{axis}
        \end{tikzpicture}
    \end{subfigure}  
    \begin{subfigure}[b]{\linewidth}
        \begin{tikzpicture}
            \begin{axis}[width=\linewidth, height=4cm, xlabel={$\Delta t_{obs}$ (ms)}, ylabel={PINAW}, xmin=0, xmax=2100, ymin=0.4, ymax=1.2, 
            xtick={0,400,...,2000}, legend pos=north east]    
            \addplot [dashed, blue, mark=diamond* ] table{Figures/FigsData/Average_PINAW500_pre.txt}; 
            \addplot [dashed, red, mark=*] table{Figures/FigsData/Average_PINAW3000_pre.txt}; 
            \end{axis}
        \end{tikzpicture}
    \end{subfigure}
\caption{Average value of PICP and PINAW metrics for predicted intervals from pre-trained models, designed for various observable post-fault durations \(\Delta t_{\text{obs}}\) and trained on different dataset sizes, evaluated across 100 test trajectories.}
    \label{fig:pre_train_results}
\end{figure}

We also performed sensitivity experiments to evaluate how different parameters affect its predictive capabilities. That is, we assess the model's performance by varying observable post-fault durations \(\Delta t_{\text{obs}}\) and training dataset sizes. Figure~\ref{fig:pre_train_results} illustrates the zero-shot prediction performance across these variables, showing that increasing both \(\Delta t_{\text{obs}}\) and dataset size improves the quality of the model's predicted intervals. In the following sections, we discuss the effects of these factors in detail.

\textit{Observable Post-Fault Duration Effect}: Figure~\ref{fig:pre_train_results} shows that increasing \(\Delta t_{\text{obs}}\) maintains the mean PICP within a comparable range. However, increasing \(\Delta t_{\text{obs}}\) up to approximately $600$ ms leads to a decrease in the value of the PINAW metric, resulting in narrower and more precise predicted intervals. Specifically, we find that PINAW decreases with respect to \(\Delta t_{\text{obs}}\) for \(\Delta t_{\text{obs}} < 600\) ms, following a power law. For instance, for the model trained on the pre-training dataset containing \(N_\mathcal{P} = 3000\) voltage trajectories, we have:
\begin{equation}
\text{PINAW} \propto \Delta t_{\text{obs}}^{-0.53}
\end{equation}
Beyond \(\Delta t_{\text{obs}} > 600\) ms, PINAW converges to a constant value, indicating that further increasing \(\Delta t_{\text{obs}}\) does not yield additional gains. This is likely due to data saturation, where the model has already captured the most significant patterns and variations in the post-fault trajectories by the time \(\Delta t_{\text{obs}}\) reaches $600$ ms. 

\textit{Dataset Size Effect}: Increasing the dataset size from 3000 to 18000 significantly improves the PICP metric while keeping the PINAW metric stable. This enhancement indicates that a larger dataset enables the model to capture a more comprehensive range of patterns and variations in post-fault trajectories, leading to more reliable predictions.

\subsubsection{Evaluating the Fine-tuned Model} \label{sec:Fine-tuned}

After pre-training the QAF-DeepONet using data from neighboring buses, we further fine-tune it with data from the target bus to enhance predictions for that specific bus.

Table~\ref{table:zero_shot_fine_tuned} presents the average metric values for the fine-tuned model's estimated predictive intervals of post-fault trajectories across 100 test cases, with the observable post-fault duration set to \(\Delta t_{obs} = 500\) ms. The corresponding pre-trained model is trained on \(N_\mathcal{P} = 6000\) voltage trajectories from neighboring buses, while the fine-tuning is conducted on \(N_\mathcal{F} = 3000\) voltage trajectories from the target bus. The results indicate that the fine-tuned model achieves higher PICP values while maintaining comparable PINAW values, demonstrating that, compared to the pre-trained model, fine-tuning enhances model performance and reliability in post-fault prediction.

\begin{table}[!h]
\centering
\begin{tabular}{cccc}
\toprule
 & \textbf{PICP\%} & \textbf{PINAW} \\
\midrule
\midrule
\textbf{Pre-trained Model} & 76.18 & 0.53  \\
\textbf{Fine tuned Model} & 81.47 & 0.51  \\
\bottomrule
\end{tabular}
\caption{Average value of PICP and PINAW metrics for estimated predictive intervals of post-fault trajectories across 100 test cases, with the observable post-fault duration set to \(\Delta t_{obs} = 500\) ms. The pre-trained model was trained on \(N_\mathcal{P} = 6000\) voltage trajectories from neighboring buses, while the fine-tuned model was refined using \(N_\mathcal{F} = 3000\) voltage trajectories from the target bus. The results demonstrate the fine-tuned model's improved performance, reflected in higher PICP values and comparable PINAW values, indicating enhanced reliability in post-fault prediction.}
\label{table:zero_shot_fine_tuned}
\end{table}

\begin{figure}[!h]
    \centering
    \begin{subfigure}[b]{\linewidth}
        \begin{tikzpicture}
            \begin{axis}[width=\linewidth, height=4cm, xlabel={$\Delta t_{obs}$ (ms)}, ylabel={PICP}, xmin=0, xmax=2100, ymin=65, ymax=95, 
            xtick={0,400,...,2000}, ytick={67,72,...,95},         
            legend style={
                at={(0.5,1.15)},
                anchor=south,
                legend columns=2, 
                column sep=1ex 
            }]    
            \addplot [dashed, blue, mark=diamond*] table{Figures/FigsData/Average_PICP500_fine.txt}; 
            \addlegendentry[font=\scriptsize]{$N_\mathcal{F} =500$}
            \addplot [dashed, red, mark=*] table{Figures/FigsData/Average_PICP3000_fine.txt}; 
            \addlegendentry[font=\scriptsize]{$N_\mathcal{F} =3000$}
            \end{axis}
        \end{tikzpicture}
        \vspace{5pt}
    \end{subfigure}  
    \begin{subfigure}[b]{\linewidth}
        \begin{tikzpicture}
            \begin{axis}[width=\linewidth, height=4cm, xlabel={$\Delta t_{obs}$ (ms)}, ylabel={PINAW}, xmin=0, xmax=2100, ymin=0.4, ymax=0.8, 
            xtick={0,400,...,2000}]    
            \addplot [dashed, blue, mark=diamond* ] table{Figures/FigsData/Average_PINAW500_fine.txt}; 
            \addplot [dashed, red, mark=*] table{Figures/FigsData/Average_PINAW3000_fine.txt}; 
            \end{axis}
        \end{tikzpicture}
    \end{subfigure}
\caption{Average value of PICP and PINAW metrics for predicted intervals from fine-tuned models, designed for various observable post-fault durations \(\Delta t_{\text{obs}}\) and trained on different dataset sizes, evaluated across 100 test trajectories. The corresponding pre-training for these models was conducted using \(N_\mathcal{P} = 6000\) voltage trajectories from neighboring buses.}
    \label{fig:fine_tune_results}
\end{figure}
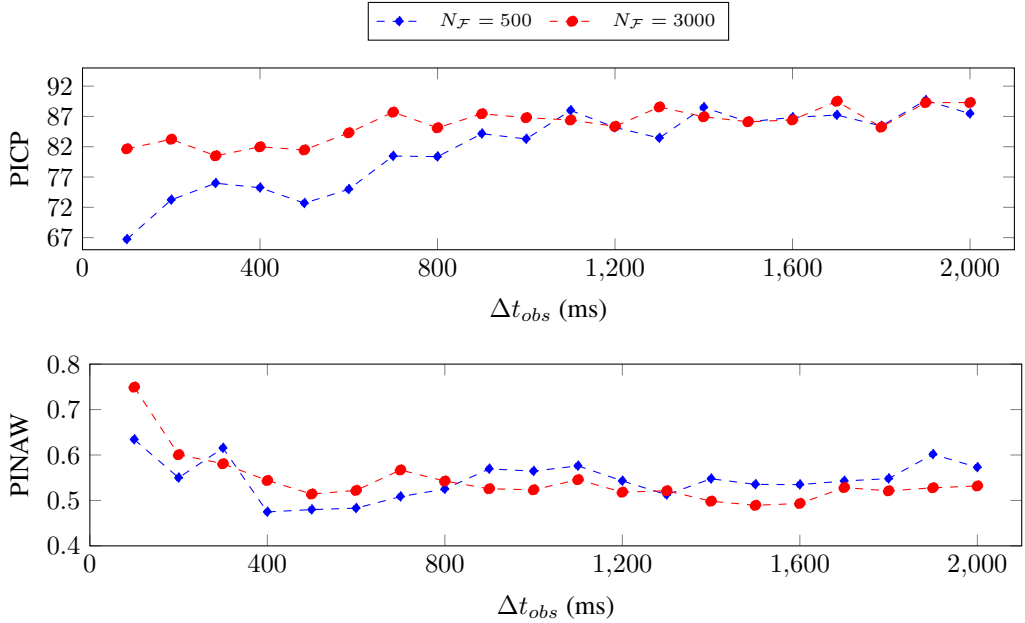

Figure~\ref{fig:fine_tune_results} illustrates the average metrics values for the fine-tuned model predictions across various \(\Delta t_{\text{obs}}\) and training dataset sizes for two different fine-tuning dataset sizes. As observed, similar to the pre-trained model, increasing the duration of observation time and the size of the training dataset enhances the model's prediction performance.

\textit{Observable Post-Fault Duration Effect}: The Figure~\ref{fig:fine_tune_results} shows that increasing $\Delta t_{obs}$ enhances both metrics. Specifically, for $\Delta t_{\text{obs}} < 1000$ ms, the value of PICP increases with  increasing observable duration, and beyond this point, it almost stabilizes. On the other hand, for $\Delta t_{\text{obs}} < 400$ ms, the value of PINAW decreases with increasing observable duration and then stabilizes. For $\Delta t_{\text{obs}} > 1000$ ms, both the PICP and PINAW values stabilize, indicating that further increases in $\Delta t_{obs}$ do not provide additional gain. This behavior is consistent with what we observed in the zero-shot prediction. 

\textit{Dataset Size Effect}: For $\Delta t_{\text{obs}} < 1000$ ms, similar to zero-shot prediction, increasing the dataset size from  \(N_\mathcal{F} = 500\)  to $3000$ in fine-tuning results in an improvement of the model's predicted intervals, reflected by a higher PICP, while maintaining the PINAW metric relatively stable. On the other hand, for $\Delta t_{\text{obs}} > 1000$ ms, the PICP metric for these datasets will almost converge. As the observation time increases, the model acquires more information about the system, reducing the amount of data required to achieve similar levels of accuracy and precision in its predictions. Consequently, for higher values of \( \Delta t_{obs} \), the PICP values converge for both dataset sizes.

\subsubsection{Evaluating Conformal Prediction} \label{sec:ConformalPrediction}

As illustrated in Figure~\ref{fig:fine_tune_results}, the average PICP values for the predicted intervals from the fine-tuned model are below the desired training goal of $95\%$. To address this, our proposed methodology integrates conformal prediction with the fine-tuned model results, modifying the predicted intervals to ensure that they achieve the desired coverage using calibration data. To this end, we calibrate the fine-tuned predictions using the calibration dataset \(\mathcal{D}_\text{cal}\), consisting of data from $N_{cal} = 100$ unseen voltage trajectories of the target bus.

Figure~\ref{fig:conforma_results} shows the average values of the metrics for the conformal prediction results. As illustrated, the predicted intervals for almost all observation times \(\Delta t_{\text{obs}}\) and both dataset sizes converge to the target value of 95\%. However, for models trained on smaller datasets and with shorter observation times, the base estimator (trained model) exhibits higher uncertainty. In such conditions, the conformal prediction PICP does not get as close to 95\% due to insufficient variability and information for accurate calibration. 

While conformal prediction adjusts the estimated predictive intervals to achieve the desired coverage, it results in increased PINAW values, leading to wider intervals. Figure~\ref{fig:conforma_results} shows that for the observable parts with \(\Delta t_{obs} < 1000\), the larger dataset results in lower PINAW values compared to the smaller dataset. This can be attributed to the larger dataset providing more comprehensive coverage of various post-fault scenarios, allowing the model to make more precise predictions. However, after \(\Delta t_{obs} > 1000\), the predictive performance of both models converges, resulting in comparable PINAW values. This convergence suggests that beyond a certain point, the benefit of having a larger dataset diminishes, and both models face similar levels of uncertainty, leading to comparable interval widths to ensure the desired coverage.

\definecolor{darkgreen}{rgb}{0.13, 0.55, 0.13}
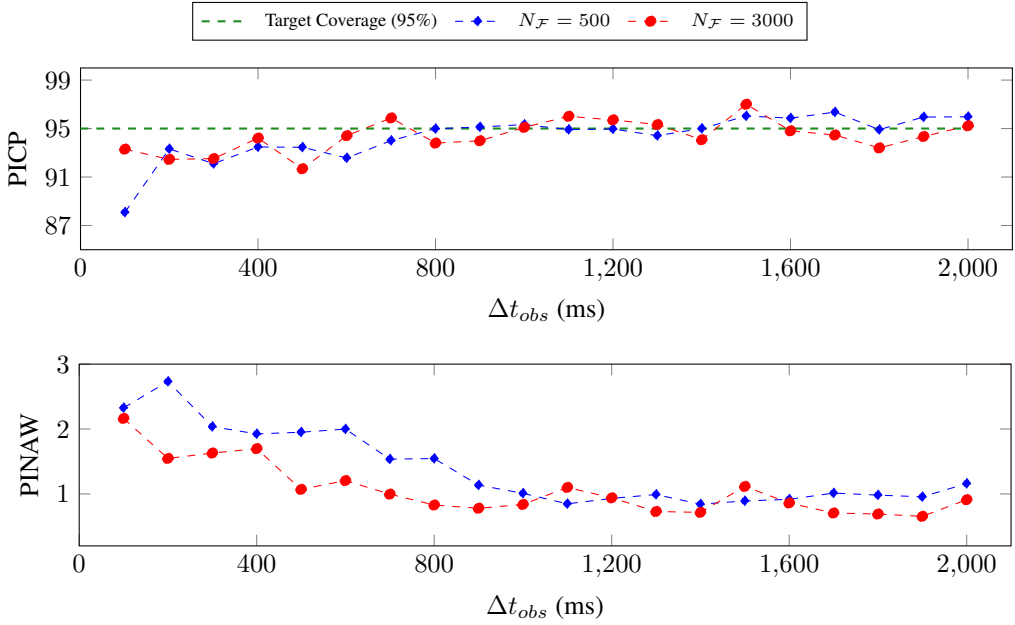
\begin{figure}[!h]
    \centering
    \begin{subfigure}[b]{\linewidth}
        \begin{tikzpicture}
            \begin{axis}[width=\linewidth, height=4cm, xlabel={$\Delta t_{obs}$ (ms)}, ylabel={PICP}, xmin=0, xmax=2100, ymin=85, ymax=100, 
            xtick={0,400,...,2000}, ytick={87,91,...,100},         
            legend style={
                at={(0.45,1.15)},
                anchor=south,
                legend columns=3, 
                column sep=1ex 
            }]   
            \addplot [dashed, darkgreen, thick] coordinates {(0,95) (2000,95)};
            \addlegendentry[font=\scriptsize]{Target Coverage (95\%)}
            \addplot [dashed, blue, mark=diamond*] table{Figures/FigsData/Average_PICP500_fine_conf.txt}; 
            \addlegendentry[font=\scriptsize]{$N_\mathcal{F} =500$}
            \addplot [dashed, red, mark=*] table{Figures/FigsData/Average_PICP3000_fine_conf.txt}; 
            \addlegendentry[font=\scriptsize]{$N_\mathcal{F} =3000$}
            \end{axis}
        \end{tikzpicture}
        \vspace{5pt}
    \end{subfigure}  
    \begin{subfigure}[b]{\linewidth}
    \hspace{2pt}
        \begin{tikzpicture}
            \begin{axis}[width=\linewidth, height=4cm, xlabel={$\Delta t_{obs}$ (ms)}, ylabel={PINAW}, xmin=0, xmax=2100, ymin=0.2, ymax=3, 
            xtick={0,400,...,2000}]    
            \addplot [dashed, blue, mark=diamond* ] table{Figures/FigsData/Average_PINAW500_fine_conf.txt}; 
            \addplot [dashed, red, mark=*] table{Figures/FigsData/Average_PINAW3000_fine_conf.txt}; 
            \end{axis}
        \end{tikzpicture}
    \end{subfigure}
    \caption{Average value of PICP and PINAW metrics for predicted intervals using the conformal prediction technique integrated with fine-tuned model results, presented across different observation post-fault durations \(\Delta t_{\text{obs}}\) and training dataset sizes. The pre-training for the fine-tuned models was conducted using  \(N_\mathcal{P} = 6000\) voltage trajectories from neighboring buses, and \({N}_{\text{cal}} = 100\) trajectories from the target bus were used to calibrate the predicted intervals of fine-tuned models within the conformal framework.} 
    \label{fig:conforma_results}
\end{figure}

As illustrated in Figure~\ref{fig:Smple_figures}, although conformal prediction uniformly extends the width of the predicted interval for both stable and unstable trajectories, the relative impact varies between the two. For stable trajectories, which generally have less variation and narrower intervals, this fixed increase results in a relatively larger expansion. In contrast, for unstable trajectories with greater variation and wider intervals, the same fixed increase results in a relatively smaller expansion. Thus, while the absolute increase in interval width is constant, its relative effect is more significant for stable trajectories and less pronounced for unstable ones. Despite this, conformal prediction still guarantees the desired coverage and enhances practical usability. Therefore, in power systems where the prediction of unstable post-fault trajectories is of high priority, incorporating conformal prediction into the approach is highly beneficial.

\subsubsection{Comparative Analysis of Results from Pre-trained Model, Fine-Tuned Model, and Conformal Prediction}

Figure~\ref{fig:Comparative Analysis} compares the pre-trained model results with those of the fine-tuned model and its corresponding conformal predictions. In this comparison, the QAF-DeepONet is first pre-trained with a dataset of \(N_\mathcal{P} = 6000\) voltage trajectories from neighboring buses and then fine-tuned using \(N_\mathcal{F} = 3000\) trajectories from the target bus. Subsequently, conformal prediction is then applied to the fine-tuned results, using \({N}_{\text{cal}} = 100\) trajectories from the target bus for calibration.

As shown in Figure~\ref{fig:Comparative Analysis}, the predicted intervals by the fine-tuned model primarily differ from the pre-trained model in their corresponding PICP values, while their PINAW values remain mostly comparable. The PICP for the fine-tuned model is considerably higher compared to zero-shot prediction, demonstrating the effectiveness of fine-tuning in improving prediction performance. This improvement indicates that fine-tuning enables the model to better capture the specific voltage dynamics of the target bus data, thereby enhancing its predictive performance. 

Comparing the results of the fine-tuned model with its conformal prediction, we observe that for almost all \(\Delta t_{\text{obs}}\), the PICP converges to the desired training goal of $95\%$, which the pre-trained and fine-tuned models do not achieve. However, conformal prediction leads to higher PINAW values compared to the fine-tuned and pre-trained models, which presents practical challenges. There is a trade-off between the guaranteed coverage probability from the conformal prediction and the width of the predicted intervals. Although conformal prediction ensures coverage and reliability, larger intervals can lead to overly conservative decisions with higher costs. Therefore, this balance should be carefully considered based on the system's objectives.

\definecolor{darkgreen}{rgb}{0.18, 0.55, 0.34}
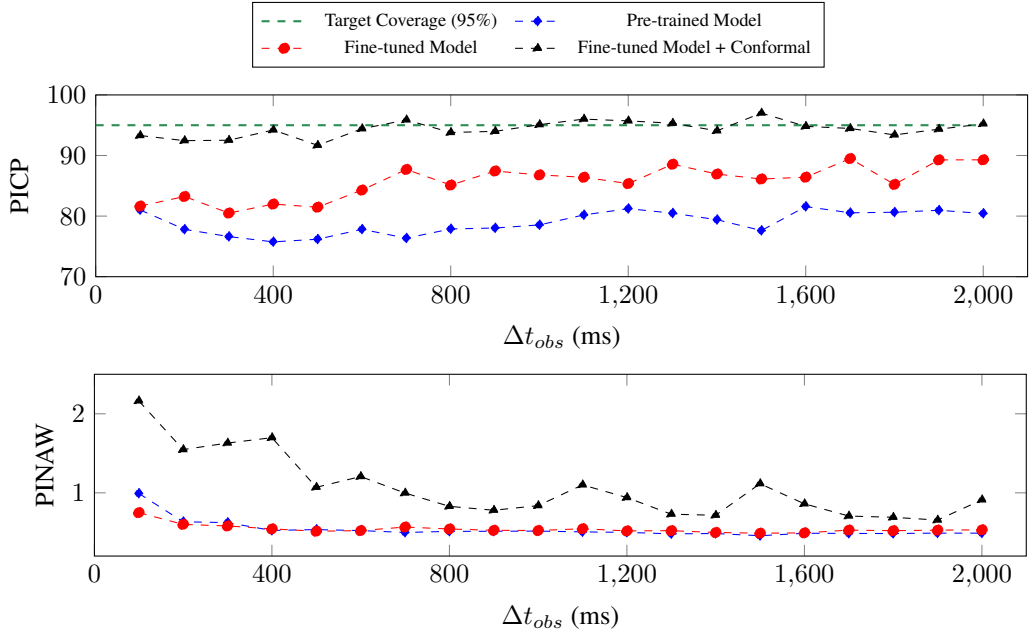
\begin{figure}[!h]
    \begin{subfigure}[b]{\linewidth}
        \begin{tikzpicture}
            \begin{axis}[width=\linewidth, height = 4cm, xlabel={$\Delta t_{obs}$ (ms)}, ylabel={PICP}, xmin=0, xmax=2100, ymin=70, ymax=100, 
            xtick={0,400,...,2000}, ytick={60,70,...,100}, 
            legend style={
                at={(0.475,1.15)},
                anchor=south,
                legend columns=2, 
                column sep=1ex 
            }] 
            
            \addplot [dashed, darkgreen, thick] coordinates {(0,95) (2000,95)};
            \addlegendentry[font=\scriptsize]{Target Coverage (95\%)}
            
            \addplot [dashed, blue, mark=diamond* ] table{Figures/FigsData/Average_PICP1000_pre.txt}; 
            \addlegendentry[font=\scriptsize]{Pre-trained Model}
            
            \addplot [dashed, red, mark=* ] table{Figures/FigsData/Average_PICP3000_fine.txt}; 
            \addlegendentry[font=\scriptsize]{Fine-tuned Model}
            
            \addplot [dashed, black, mark=triangle* ] table{Figures/FigsData/Average_PICP3000_fine_conf.txt}; 
            \addlegendentry[font=\scriptsize]{Fine-tuned Model + Conformal }
            \end{axis}
        \end{tikzpicture}
        \vspace{5pt}
    \end{subfigure}  
    \begin{subfigure}[b]{\linewidth}
        \hspace{7pt}
        \begin{tikzpicture}
            \begin{axis}[width=\linewidth, height = 4cm, xlabel={$\Delta t_{obs}$ (ms)}, ylabel={PINAW}, xmin=0, xmax=2100, ymin=0.2, ymax=2.5, 
            xtick={0,400,...,2000}] 
   
            \addplot [dashed, blue, mark=diamond* ] table{Figures/FigsData/Average_PINAW1000_pre.txt}; 
            
            \addplot [dashed, red, mark=* ] table{Figures/FigsData/Average_PINAW3000_fine.txt}; 
            
            \addplot [dashed, black, mark=triangle* ] table{Figures/FigsData/Average_PINAW3000_fine_conf.txt}; 
            \end{axis}
        \end{tikzpicture}
    \end{subfigure}
    \caption{Comparative analysis of predicted intervals by the pre-trained model, fine-tuned model, and conformal prediction, showing differences in PICP and PINAW average values across various \(\Delta t_{\text{obs}}\). The QAF-DeepONet was pre-trained with  \(N_\mathcal{P} = 6000\) voltage trajectories from neighboring buses and fine-tuned with  \(N_\mathcal{F} = 3000\) trajectories from the target bus. Conformal prediction was then applied to the fine-tuned results, using \({N}_{\text{cal}} = 100\) trajectories from the target bus for calibration}

    \label{fig:Comparative Analysis}
\end{figure}
\section{Future Work} \label{sec:future-work}

The conformal prediction method guarantees that the PICP converges to the desired value. However, this method does not address the PINAW metric, resulting in wider prediction intervals during the calibration process. PINAW is a crucial metric for effective decision-making as it directly impacts the practical usability of the prediction intervals.

To address this limitation, our future work will incorporate conformal risk control\citep{angelopoulos2022conformal}. This enhanced approach will not only ensure the convergence of PICP to the desired goal but will also consider and optimize PINAW. By doing so, we aim to develop a more comprehensive methodology for estimating predictive intervals of post-fault trajectories that balances coverage probability and interval width, thereby improving its practical applicability in decision-making.

In addition, this work represents a key step in developing foundation models~\citep{bommasani2021opportunities} for transient stability. Moving forward, we aim to design merging and federated pre-training strategies for transformers~\citep{vaswani2017attention} that enable creating larger, more capable foundation models that can effectively forecast trajectories of complex power system responses.
\section{Conclusion} \label{sec:conclusion}
This paper presents a novel methodology for estimating predictive intervals for post-fault voltage trajectories in power systems using deep operator learning techniques. We introduce the Quantile Attention-Fourier Deep Operator Network (QAF-DeepONet), which combines deep operator networks with attention mechanisms and Fourier features to handle complex voltage trajectory patterns. QAF-DeepONet maps the observed segments of the voltage trajectory to predictive intervals for the post-fault voltage trajectory. To address the low-regime data issue in power systems, the methodology incorporates a two-stage training approach. We initially pre-train the QAF-DeepONet with data from neighboring buses to learn the underlying dynamics within voltage trajectories. This approach makes maximum use of available data and ensures each bus's data privacy through a federated learning method that avoids direct data sharing. Subsequently, we fine-tune the model using data from the target bus to learn the specific characteristics of the dynamics of its voltage trajectories. Next, we integrate conformal prediction with the fine-tuned model's predictions to provide coverage guarantees. We evaluated our approach using data from the IEEE-39 bus test system through a series of comprehensive analyzes. 

The results underscore the efficacy of the proposed methodology, demonstrating precise and reliable performance in zero-shot predictions for estimating the predictive interval. Furthermore, the results demonstrate that fine-tuning improves the reliability and robustness of the model predictions. Additionally, the integration of conformal prediction techniques enhances the model's reliability by ensuring that the predicted intervals achieve the target coverage probability. However, this integration also tends to result in wider intervals, illustrating the inherent trade-off between reliability and precision.
\section*{Acknowledgement}
G. Lin would like to thank the support by National Science Foundation (DMS-2053746, DMS-2134209, ECCS-2328241, CBET-2347401 and OAC-2311848), and U.S.~Department of Energy (DOE) Office of Science Advanced Scientific Computing Research program DE-SC0023161, and DOE–Fusion Energy Science, under grant number: DE-SC0024583. C. Li would like to thank the support of the startup funding from the Davidson School of Chemical Engineering and the College of Engineering at Purdue University.
\bibliographystyle{plainnat}
\bibliography{refpaper.bib}

\end{document}